%% file: main.tex
\documentclass[10pt,twocolumn,letterpaper]{article}
\usepackage[dvipsnames,table]{xcolor}
\usepackage{cvpr}              % To produce the CAMERA-READY version

\definecolor{cvprblue}{rgb}{0.21,0.49,0.74}
\usepackage[pagebackref,breaklinks,colorlinks,allcolors=cvprblue]{hyperref}
\usepackage{multirow}
\usepackage{adjustbox}
\usepackage{bm}
\usepackage{pifont}
\usepackage{makecell}
\usepackage{ulem}
\usepackage[accsupp]{axessibility}
\def\paperID{*****} % *** Enter the Paper ID here
\def\confName{CVPR}
\def\confYear{2025}

\newcommand{\mypara}[1]{\smallskip\noindent\textbf{#1}}
\title{DeCLIP: Decoupled Learning for Open-Vocabulary Dense Perception}

\author{
    Junjie Wang\textsuperscript{1} \quad
    Bin Chen\textsuperscript{2,3,\thanks{Corresponding authors}} \quad
    Yulin Li\textsuperscript{1} \quad
    Bin Kang\textsuperscript{3} \quad
    Yichi Chen\textsuperscript{3} \quad
    Zhuotao Tian\textsuperscript{1,\footnotemark[1] } \\
    \textsuperscript{1}School of Computer Science and Technology, HIT, Shenzhen \\
    \textsuperscript{2}International Research Institute for Artificial Intelligence, HIT, Shenzhen \\
    \textsuperscript{3}University of Chinese Academy of Sciences \\
}
\begin{document}
\maketitle
\input{0_abstract}    
\input{1_intro}
\input{1_5_preliminary}
\input{2_method}

\input{3_experiment}
\input{5_conclusion}
{
    \small
    \bibliographystyle{ieeenat_fullname}
    \bibliography{main}
}
\input{6_suppl}


\end{document}

%% file: 0_abstract.tex
\begin{abstract}
% Contrastive Language-Image Pre-training (CLIP) has significantly advanced Open-Vocabulary Dense Prediction tasks through its zero-shot capabilities. However, since CLIP was initially trained for image-level alignment, directly applying it to downstream dense prediction tasks such as object detection and image segmentation often leads to suboptimal performance. In this paper, we conduct an in-depth analysis on the attention maps of CLIP’s [CLS] token and image tokens, revealing that attention interference from the [CLS] to image tokens contributes to this performance degradation. To address this issue, we propose DeCLIP, a novel fine-tuning framework that enhances the discriminability and spatial consistency of CLIP’s local features via a decoupled feature enhancement strategy. Extensive experiments demonstrate that DeCLIP can be effectively applied to mainstream open-vocabulary dense prediction tasks, outperforming state-of-the-art methods across a broad range of benchmarks. The code will be made publicly available.
Dense visual prediction tasks have been constrained by their reliance on predefined categories, limiting their applicability in real-world scenarios where visual concepts are unbounded. While Vision-Language Models (VLMs) like CLIP have shown promise in open-vocabulary tasks, their direct application to dense prediction often leads to suboptimal performance due to limitations in local feature representation. In this work, we present our observation that CLIP's image tokens struggle to effectively aggregate information from spatially or semantically related regions, resulting in features that lack local discriminability and spatial consistency. To address this issue, we propose DeCLIP, a novel framework that enhances CLIP by decoupling the self-attention module to obtain ``content'' and ``context'' features respectively. The ``content'' features are aligned with image crop representations to improve local discriminability, while ``context'' features learn to retain the spatial correlations under the guidance of vision foundation models, such as DINO. Extensive experiments demonstrate that DeCLIP significantly outperforms existing methods across multiple open-vocabulary dense prediction tasks, including object detection and semantic segmentation. Code is available at \textcolor{magenta}{https://github.com/xiaomoguhz/DeCLIP}.
\end{abstract}

%% file: 1_intro.tex
\section{Introduction}
\label{sec:intro}
In the era of deep learning, dense prediction tasks like object detection \cite{fasterrcnn,dab_detr} and image segmentation \cite{unet,mask2former} have rapidly advanced and are widely used. However, traditional methods \cite{maskdino,sqr_detr,dedetr} recognize only a fixed set of predefined categories. This restriction hinders the practical application of these methods in real-world settings, where the range of visual concepts is virtually boundless. Consequently, increasing attention has been drawn to open-vocabulary methods \cite{ovr-cnn,wu2023aligning,wu2023cora,catseg}, which aim to detect and segment objects from any category using textual descriptions.
\begin{figure}[tbp]
  \centering
  \includegraphics[width=\linewidth]{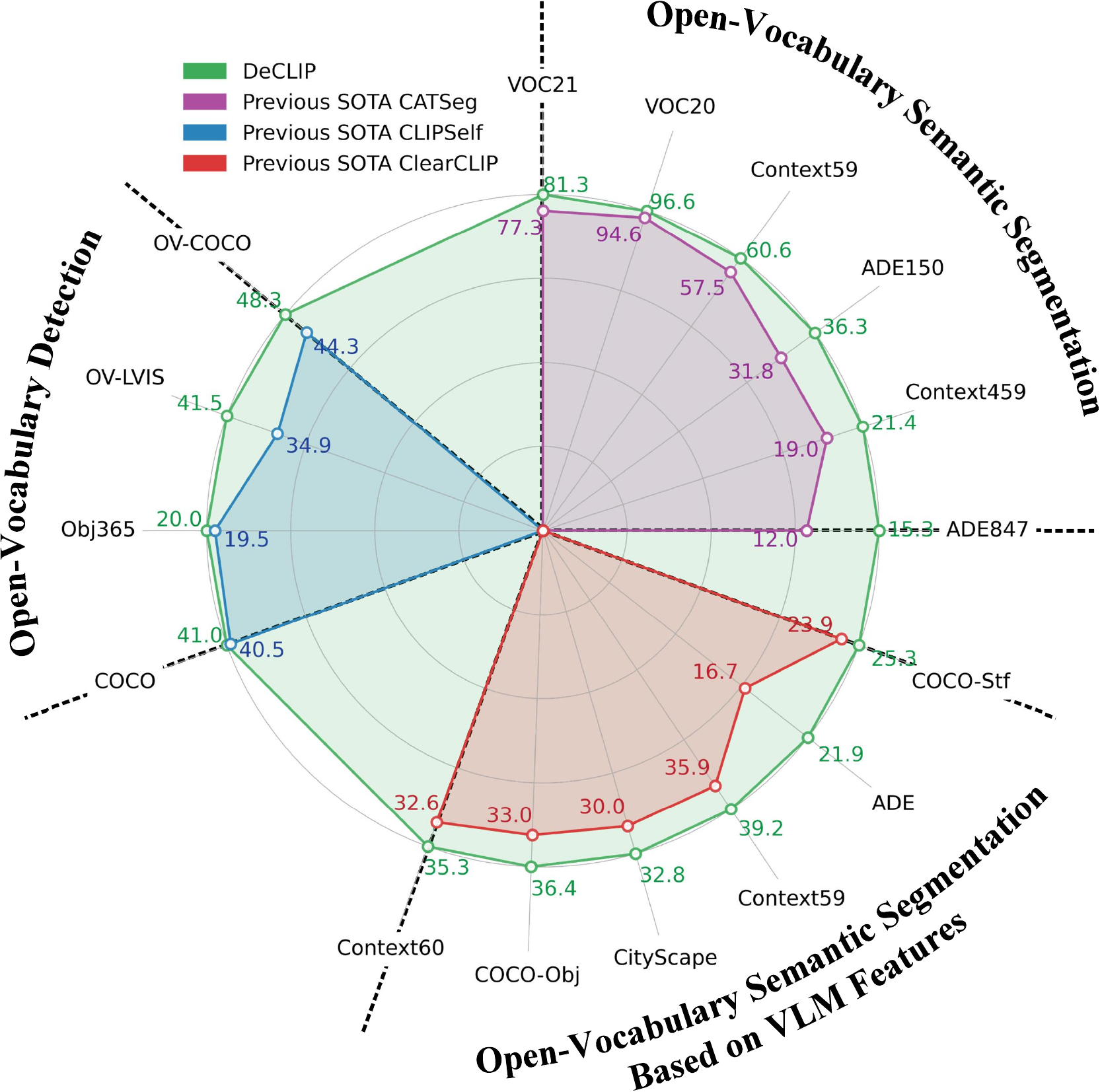}
  \caption{DeCLIP outperforms previous state-of-the-art models on a broad range of open-vocabulary dense prediction benchmarks.}
  \label{fig1}
\end{figure}
\begin{figure*}[htbp]
\centering
\includegraphics[width=0.85\linewidth]{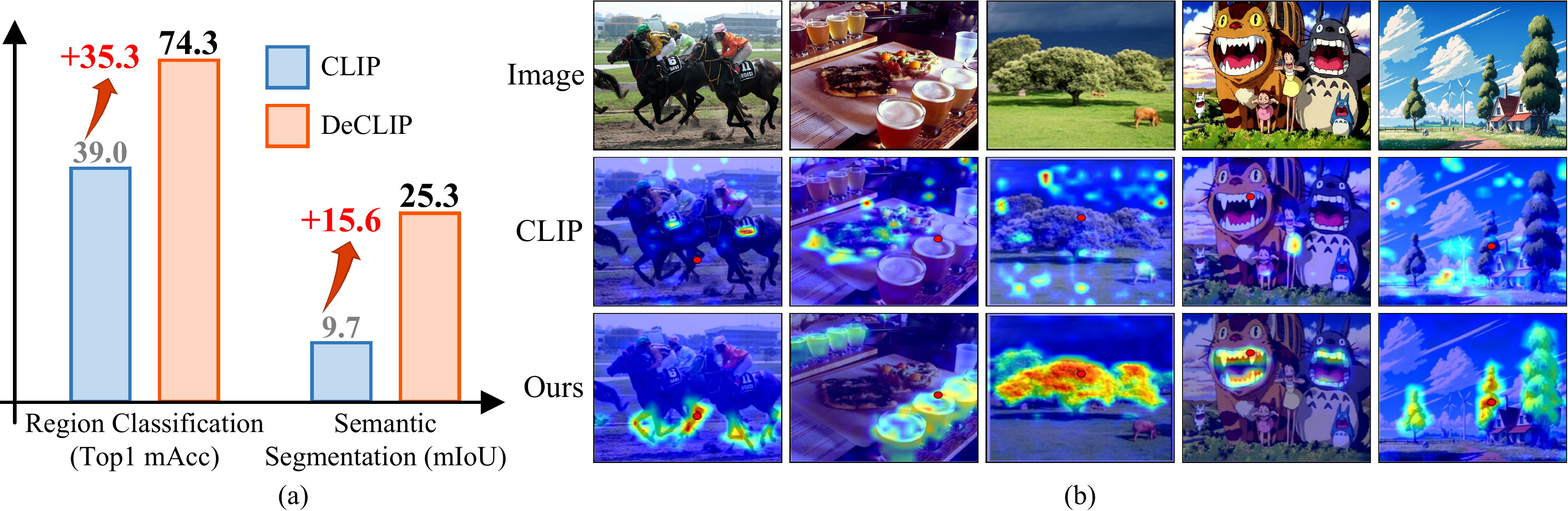}
\caption{\textbf{Quantitative and qualitative comparisons between our method and CLIP.} \textbf{(a)} Performance comparisons of open-vocabulary dense predictions on COCO \cite{mscoco}. \textbf{(b)} Attention map comparisons, with the anchor image token marked in red.}

\label{fig2}
\end{figure*}
\par Building on the success of Vision-Language Models (VLMs) \cite{clip,evaclip,openclip,flip} pre-trained on image-text pairs, such as CLIP \cite{clip}, researchers have started leveraging these models for open-vocabulary dense prediction tasks. Among these \cite{wu2023aligning,clim,wu2023clipself,sasdet,ovdquo,rtgen}, transfer-learning approaches \cite{frozenseg,ovdquo,fvlm,fcclip,wu2023clipself,MAFT} have shown outstanding performance. These methods utilize the image encoder of VLM as a feature extractor and exclusively train lightweight task-specific components. Whereas using VLMs as feature extractors offers significant advantages due to their comprehensive pre-training, directly applying these image-level models to dense prediction often leads to domain shift issues \cite{wu2023cora,wu2023clipself}.

\par
\begin{figure*}[htbp]
\centering
\includegraphics[width=0.9\linewidth]{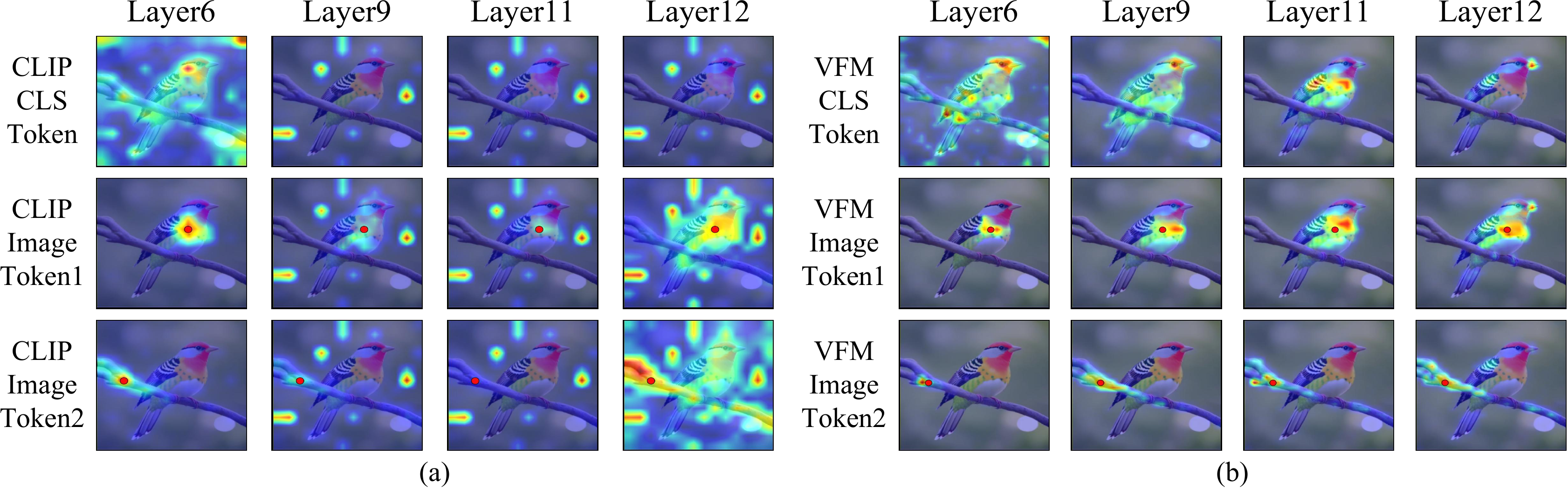}
\caption{\textbf{Visualization of attention maps across different encoding layers of CLIP and VFM.} The attention weights are calculated at a low resolution, then averaged across different heads, and finally upsampled to the original image resolution for visualization. The anchor image token is marked in red. We observe the occurrence of the ``proxy'' token phenomenon in CLIP, but not in VFM. Furthermore, when the position of the anchor image token is shifted, VFM shows a better correlation for image tokens with the same semantics.}
\label{fig3}
\end{figure*}

 \mypara{What hinders CLIP in dense perception? }
To assess VLM's constraints in dense perception, we analyze CLIP's attention maps across various layers (Figure \ref{fig3}(a)). Our experiments reveal that CLIP's [CLS] token may interfere with the correlations among other image tokens, leading to suboptimal performance in dense prediction tasks. 

Specifically, we have observed that in deeper layers (behind the 9th layer), the [CLS] token shifts focus away from primary objects within the image and attends highly to certain background tokens, as highlighted by the bright spots in the first row of Figure \ref{fig3}(a). Moreover, image tokens (rows 2 and 3, Figure \ref{fig3}(a)) exhibit similar behavior with the [CLS] token, showing high attention to certain background tokens regardless of their positions.
\par This observation sheds light on why CLIP struggles in dense prediction tasks: its image tokens fail to aggregate information from spatially or semantically related regions, resulting in dense features that lack local discriminability and spatial consistency. As shown in Figure \ref{fig2}(a), directly using CLIP features on the COCO dataset yields relatively inferior performance in open-vocabulary region classification and semantic segmentation. 
To tackle this, an intuitive approach is to enhance CLIP's local representations through fine-tuning. However, balancing the optimizations of both local feature spatial correlations and vision-language semantic alignment within a unified architecture becomes a new challenge. Therefore, \textit{is it feasible to disentangle CLIP’s features and apply separate guiding constraints to obtain diverse features within a unified architecture?}

\par \mypara{Our solution. }
To address these challenges, we propose DeCLIP, a general unsupervised fine-tuning method aimed at enhancing both the discriminability and spatial consistency of CLIP’s local features. The core idea is to decouple the self-attention module of CLIP and learn from different teacher models separately.

Specifically, DeCLIP decouples the features in the self-attention module into “\textit{content}” and “\textit{context}” components. The “content” features, responsible for local discriminability, are fine-tuned by aligning pooled region features with their corresponding image crop [CLS] representations. Meanwhile, the “context” features, responsible for spatial consistency, are learned from the feature correlations generated by Vision Foundation Models (VFMs). This decoupled distillation design effectively mitigates optimization conflicts, improving the generalization ability when applying CLIP to downstream open-vocabulary dense prediction tasks. As shown in Figure \ref{fig2}, DeCLIP significantly outperforms CLIP in local discriminability and spatial consistency.
\begin{figure*}[htbp]
\centering
\includegraphics[width=0.9\linewidth]{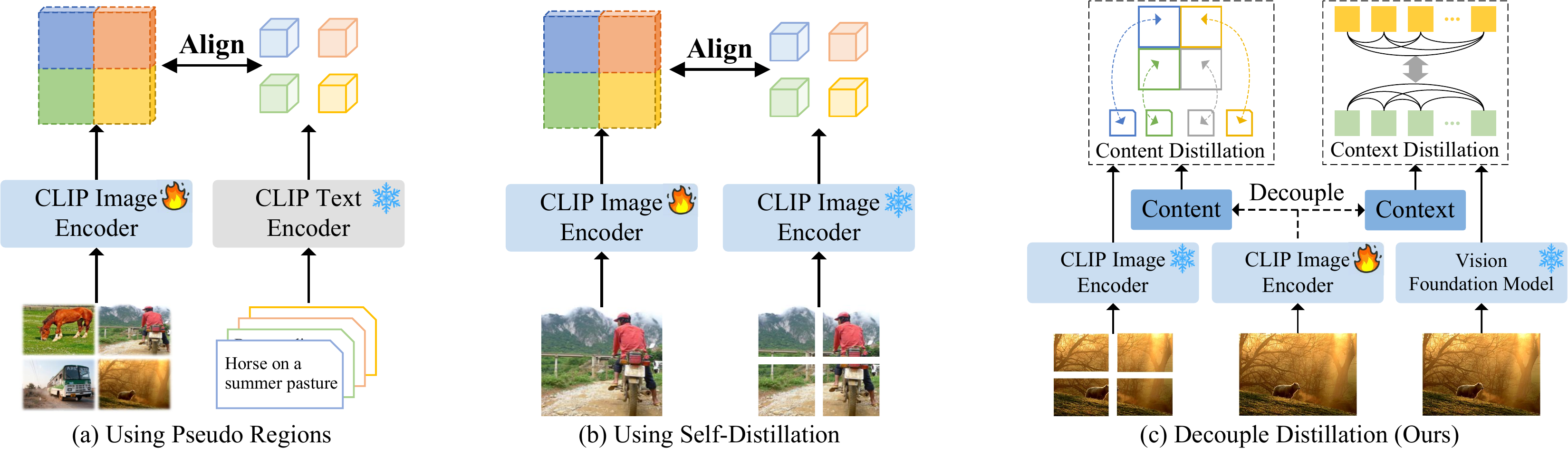}
\caption{\textbf{Pre-fine-tuning methods for adapting CLIP to dense prediction tasks.} Existing work considers establishing region-text alignment through cost-effective methods via: \textbf{(a)} using images as pseudo regions or \textbf{(b)} using self-distillation on image patches. The former regards the entire image as a region, which results in a loss of details. The latter uses self-distillation on the image patches thereby gaining more fine-grained information, but still fails to apply to pixel-level image segmentation. \textbf{(c)} Unlike prior approaches, we use VFM to guide the spatial consistency of CLIP's features, and decouple CLIP's features for distillation separately to avoid optimization conflicts.}
\label{fig4}
\end{figure*}
To summarize, our contributions are as follows:

\begin{itemize}
\item We analyze CLIP and find that its limitation in open-vocabulary dense prediction arises from image tokens failing to aggregate information from spatially or semantically related regions.
\item To address this issue, we propose DeCLIP, a simple yet effective unsupervised fine-tuning framework, to enhance the discriminability and spatial consistency of CLIP's local features via a decoupled feature enhancement strategy.
\item Extensive experiments demonstrate that DeCLIP can be decently applied to mainstream open-vocabulary dense prediction tasks, including object detection and semantic segmentation. As illustrated in Figure~\ref{fig1}, DeCLIP outperforms state-of-the-art methods across a broad range of benchmarks, achieving superior performance metrics in all evaluated task domains.

\end{itemize}

%% file: 1_5_preliminary.tex
\section{Background and Motivation
}
In the following, we provide a concise overview of foundational concepts pertinent to this study in Section~\ref{preliminary}, and highlight important findings in Section~\ref{observation}, which offer valuable insights for motivating the proposed approach. 
% The detailed related work is in the appendix.

\subsection{Preliminaries}
\label{preliminary}
\mypara{Contrastive Language-Image Pre-training} (CLIP) \cite{clip} is built upon two encoders, one for images and one for text. The visual encoder of CLIP can be a CNN series \cite{convnext,resnet} or ViT \cite{ViT}, and the text encoder is a Transformer \cite{transformer}. This paper focuses on the CLIP model with the ViT architecture, which adopts the [CLS] token to represent the overall features of an image. CLIP learns vision-language alignment by maximizing the cosine similarity between the [CLS] token and text features of matched image-text pairs, and minimizing the similarity for unmatched pairs.

\mypara{Dense feature extraction with CLIP.} ViT-based CLIP consists of a series of stacked attention blocks. For example, the ViT-B version of CLIP includes 12 attention block layers. Let $\mathbf{X}=\{\bm{x}_0, \bm{x}_1, \cdots, \bm{x}_{h \times w}\}$ denotes the input to the last attention block, where $\bm{x}_i \in \mathbb{R}^{1 \times D}$. The computation within this attention block can be expressed as:
\begin{align}
\mathbf{Q} &= \text{Proj}_q(\mathbf{X}), \, \mathbf{K} = \text{Proj}_k(\mathbf{X}), \, \mathbf{V} = \text{Proj}_v(\mathbf{X}), \\
\mathbf{Y} &= \mathbf{X} + \text{Proj}\left(\text{Attn}_{qk} \cdot \mathbf{V}\right), \\
\mathbf{Z} &= \mathbf{Y} + \text{FFN}(\mathbf{Y}),
\end{align}
where $\mathbf{Q}$, $\mathbf{K}$, and $\mathbf{V}$ represent the query, key, and value embeddings, respectively; $\text{Proj}$ denotes projection layers; $\text{Attn}_{qk}=\text{SoftMax}\left(\mathbf{Q}\mathbf{K}^\top / \sqrt{d}\right)$ represents the self-attention process, with $d$ denoting the dimension of each attention head. $\text{FFN}$ denotes a feed-forward network. For simplicity, normalization operations are omitted. 

After passing through the final attention block, $\mathbf{Z}[0]$ represents the global [CLS] token. The remaining image patch embeddings $\mathbf{Z}[1:h \times w]$ can be reshaped to obtain dense feature representations $\mathbf{X}_{\text{dense}} \in \mathbb{R}^{C \times H \times W}$\footnote{The final V-L projection layer is omitted here for brevity.}.
\par \mypara{Adapting CLIP to dense prediction tasks.} Several studies have attempted to alleviate the domain shift issue in applying CLIP to dense prediction tasks via fine-tuning strategies. These approaches fall into two main categories:
\begin{itemize}
\item \mypara{Joint fine-tuning.} These methods fine-tune CLIP while training task-specific components \cite{MAFT,catseg,lseg,OVSeg,maskqclip,maft+,xie2024sed}. For instance, CAT-Seg \cite{catseg} proposes an attention fine-tuning strategy based on ViT CLIP, which generalizes well to unseen categories. MAFT \cite{MAFT} leverages attention bias to fine-tune CLIP for mask classification. 
\item \mypara{Pre-fine-tuning.} These methods directly fine-tune CLIP using cost-efficient techniques \cite{wu2023clipself,pacl,regionclip,clim,wu2023cora}, which are more closely aligned with the approach proposed in this paper. As illustrated in Figure \ref{fig4}(a), CLIM \cite{clim} employs a mosaic augmentation technique to stitch multiple images into a single image, enabling each sub-image to serve as a pseudo-region for region-text contrastive learning. CLIPSelf \cite{wu2023clipself} enhances CLIP's region classification accuracy by maximizing cosine similarity between its region representations and the corresponding image crop representations, as illustrated in Figure \ref{fig4}(b).
\end{itemize}

\subsection{Key Observations}
\label{observation}
Despite the promising results of the two categories of fine-tuned methods in Section \ref{preliminary}, they continue to exhibit certain limitations. Joint fine-tuning methods are typically specific to tasks or models and heavily rely on labor-intensive annotations of dense prediction tasks. On the other hand, pre-fine-tuning methods demonstrate broader applicability. However, their region-level fine-tuning technique remains limited in image segmentation tasks that require pixel-level details.
To tackle this issue, we investigate the feasibility of incorporating pixel-level details into CLIP's pre-fine-tuning, enabling it to better align with open-vocabulary dense prediction tasks. In the following, we start by analyzing CLIP’s attention maps across various layers.

\mypara{The ``proxy'' token phenomenon.} As shown in Figure \ref{fig3}(a), we found that in CLIP's shallow layer, the attention weights of CLIP’s [CLS] token are widely distributed across the image (i.e., layer 6). However, in the deeper layers, the [CLS] token shifts its focus away from primary objects in the image and attends to specific tokens, as highlighted by the bright spots within the image background. Additionally, we found that image tokens (rows 2 and 3) exhibit similar behavior to the [CLS] token, showing high attention to certain tokens in the background, regardless of their position. 

These background tokens may serve as ``proxies'' for the [CLS] token. This suggests that these tokens aggregate essential information from other image tokens, enabling the [CLS] token to form an approximate ``global view'' by summarizing content from them, thereby facilitating image classification. However, these ``proxy'' tokens negatively affect the feature correlations between image tokens. As illustrated in Figure \ref{fig3}(a), when we shift the position of the anchor image token (from the bird to the branch), we observe that the new image token still pays high attention to the ``proxy'' tokens. This results in a lack of correlation between image patches that share the same semantics, which is detrimental to dense prediction tasks.

 \mypara{VFMs exhibit better dense correlations. }
Considering the inherent constraints that impede CLIP's efficacy in dense perception tasks, we instead observe that VFMs such as the DINO series \cite{dino,dinov2}, trained in a self-supervised manner, and the SAM series \cite{sam,sam2}, trained on large-scale segmentation data, are capable of extracting features with strong spatial consistency, as shown in Figure \ref{fig3}(b).

 \begin{table}[tbp]
    \centering
    \caption{Performance of different distillation schemes.}
    \begin{adjustbox}{width=\linewidth,center,valign=t}
    \begin{tabular}{l|cccc}
        \toprule
        \multirow{2.5}{*}{Distillation Type} & \multicolumn{2}{c}{Region Classification (mAcc)} & \multicolumn{2}{c}{Semantic Segmentation (mIoU)} \\
        \cmidrule(lr){2-3} \cmidrule(lr){4-5}
        & COCO (Thing) & COCO (Stuff) & Context59 & CityScape \\
        \midrule
        Self Distillation \cite{wu2023clipself} & 69.5 & 44.6 & 29.4 & 25.6 \\
        Self+VFM Distillation \cite{sam} & 65.6 \textcolor[HTML]{ff0000}{(-3.9)} & 41.3 \textcolor[HTML]{ff0000}{(-3.3)} & 32.4 \textcolor[HTML]{159838}{(+3.0)} & 28.7 \textcolor[HTML]{159838}{(+3.1)} \\
        Self+VFM+Decouple & 75.0 \textcolor[HTML]{159838}{(+5.5)} & 51.8 \textcolor[HTML]{159838}{(+7.2)} & 35.3 \textcolor[HTML]{159838}{(+5.9)} & 32.3 \textcolor[HTML]{159838}{(+6.7)} \\
        \bottomrule
    \end{tabular}
    \end{adjustbox}
\label{sanity}
\end{table}
 
 In particular, the attention map of VFMs does not exhibit the ``proxy" token phenomenon observed in CLIP. Furthermore, when we change the position of the anchor image token, the VFM shows a better correlation for image tokens with the same semantics. Therefore, we consider whether VFMs can be incorporated into the pre-fine-tuning process to further improve the feature correlations of CLIP. However, this straightforward approach fails to achieve satisfactory results. Conducting VFM distillation\footnote{``VFM distillation'' indicates aligning the feature self-correlations between CLIP's $\mathbf{X}_{\text{dense}}$ and that of the VFM.} and self-distillation\footnote{``Self-distillation'' refers to aligning region features from $\mathbf{X}_\text{dense}$ with their corresponding [CLS] representation.} simultaneously results in reduced region classification performance, as shown in Table \ref{sanity} (row 2). We hypothesize that this observation stems from the fact that spatial feature correlation and vision-language alignment have different optimization focuses, and optimizing them simultaneously within a single model results in trade-offs.
 

%% file: 2_method.tex
\begin{figure*}[htbp]
\centering
\includegraphics[width=0.9\linewidth]{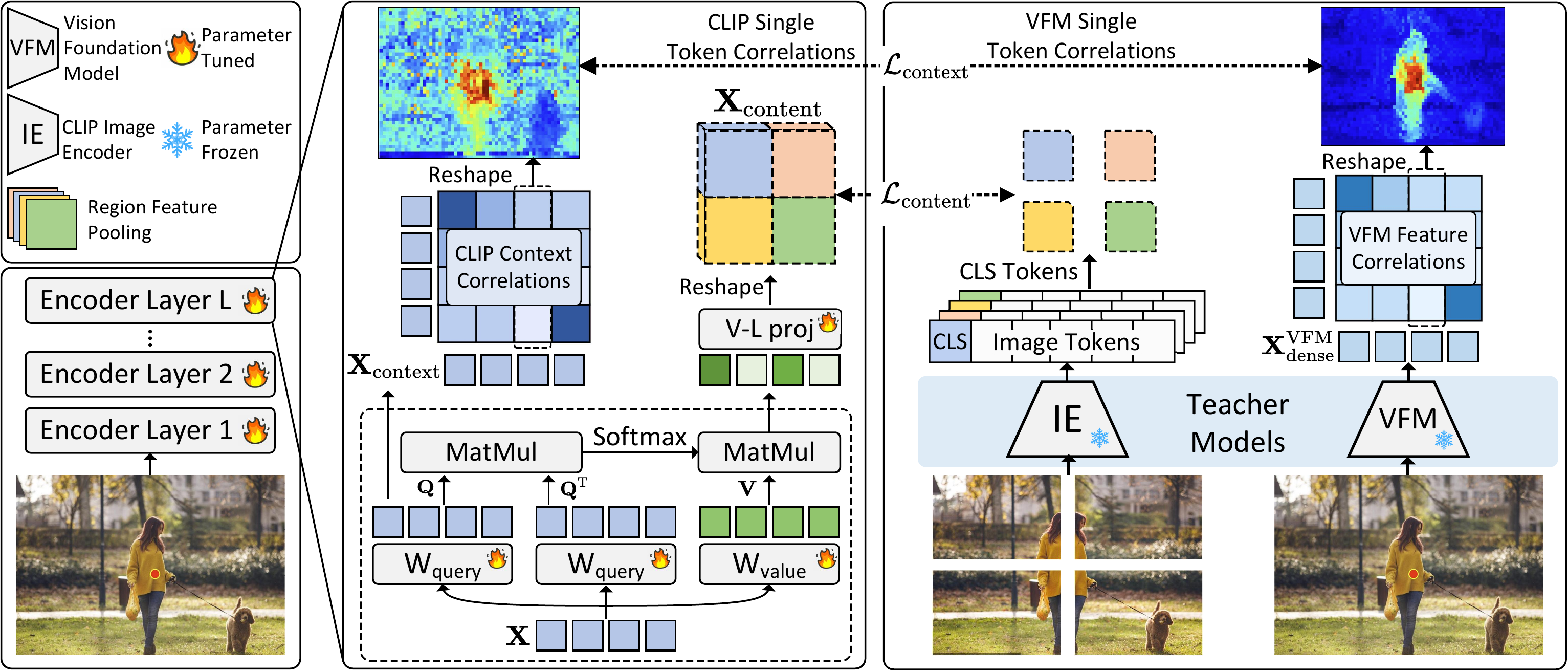}
% \caption{\textbf{Illustration of the DeCLIP framework.} We decouple CLIP's final attention module into context and content features for distillation, avoiding optimization conflicts between feature correlations and visual-language alignment. The teacher for the content feature is CLIP itself, improving the region classification accuracy. The teacher for the context feature is a VFM, aimed at enhancing the spatial consistency of features. }

\caption{\textbf{Illustration of the DeCLIP framework.} We decouple CLIP's final attention module into context and content features for distillation, avoiding optimization conflicts between feature correlations and visual-language alignment. CLIP itself serves as the teacher for content features to improve region classification accuracy. A VFM serves as the teacher for context features to enhance spatial consistency. }
\label{fig5}
\end{figure*}
\section{Method}
Through the above analysis, we found that CLIP underperforms in dense prediction tasks since its image tokens fail to effectively aggregate information from semantically related regions. Observations of VFMs' attention maps inspired us to incorporate them into CLIP's pre-fine-tuning process. Considering the optimization conflict between feature correlations and visual-language alignment, we applied a decoupled feature enhancement strategy to CLIP.

\par In this section, we introduce DeCLIP, an unsupervised fine-tuning framework for adapting CLIP to dense prediction tasks. We first explain how to decouple CLIP’s self-attention mechanism into “content” and “context” components in Sec.\ref{sec3.1}, then describe how these components learn from different ``teacher'' models in Sec.\ref{declip} by distillation.

\subsection{Decoupled Attention}
\label{sec3.1}
 The unsuccessful attempts to simultaneously perform self-distillation and VFM distillation on $\mathbf{X}_{\text{dense}}$ (Table \ref{sanity}, row 2) prompted us to explore the feasibility of a decoupled distillation. In the following, we propose decoupling CLIP's self-attention module to obtain “content” and “context” features, and separately optimize the local discriminability and spatial consistency abilities, as illustrated in Figure \ref{fig4}(c).

\mypara{Rethinking the self-attention.} As described in Sec.\ref{preliminary}, in CLIP’s last attention block, the $\mathbf{V}$ features are weighted and summed under the guidance of the attention map ($\text{Attn}_{qk}$) derived from $\mathbf{Q}$ and $\mathbf{K}$, which define spatial or semantic relationships among image tokens. Studies \cite{sclip,CLIPtrase,clearclip,clipdino} have shown that CLIP's dense features $\mathbf{X}_{\text{dense}}$ can be directly used for semantic segmentation by per-pixel classification, indicating that each pixel of $\mathbf{X}_{\text{dense}}$ contains independent semantic information. Inspired by this, we regard $\mathbf{Q}$ and $\mathbf{K}$ as anchors for improving spatial consistency, and $\mathbf{X}_{\text{dense}}$ as an anchor for enhancing local discriminability.

\par Additionally, recent training-free OVS studies \cite{sclip,clearclip} have further promoted us to decouple CLIP’s self-attention followed by distillation. They modify CLIP’s attention block from $\text{Attn}_{qk}$ to $\text{Attn}_{qq}$ and remove the residual connections, simplifying the optimization of local feature consistency by focusing on $\mathbf{Q}$ alone. Based on our rethinking of CLIP's self-attention and inspired by these methods, we propose decoupling CLIP's last attention block to obtain “content” and “context” features for distillation as follows:
\begin{align}
&\mathbf{X}_{\text{context}} = \text{Proj}_q(\mathbf{X}), \,  \mathbf{V} = \text{Proj}_v(\mathbf{X}), \\
&\mathbf{X}_{\text{content}} =\text{Proj}\left(\text{Attn}_\text{context} \cdot \mathbf{V}\right),
\\
&\text{Attn}_\text{{context}}=\text{SoftMax}\left(\mathbf{X}_{\text{context}}\mathbf{X}_{\text{context}}^\top / \sqrt{d}\right).
\label{eq5}
\end{align}
\par Specifically, $\mathbf{V}$ is aggregated based on the attention map ($\text{Attn}_\text{context}$) generated from $\mathbf{X}_{\text{context}}$. $\mathbf{X}_{\text{context}}$ determines which image tokens are semantically or spatially related. $\mathbf{X}_{\text{content}}$ carries the semantic information of each image token in the visual-language space. By decoupling the features in this manner, we can apply different guidance constraints to $\mathbf{X}_{\text{context}}$ and $\mathbf{X}_{\text{content}}$ to obtain diverse feature representations in a unified architecture without interference. 

As observed in Sec.~\ref{observation}, VFM exhibits a strong correlation for image tokens with the same semantics, thus we leverage it as guidance for $\mathbf{X}_{\text{context}}$ to improve CLIP's local feature spatial consistency. Meanwhile, we employ the self-distillation technique as guidance for $\mathbf{X}_{\text{content}}$ to enhance the visual-language alignment of CLIP's region feature.

\par As demonstrated in Table \ref{sanity} (row 3), this decoupled optimization significantly improves the local discriminability and spatial consistency of CLIP’s features, leading to simultaneous enhancements in both region classification accuracy and semantic segmentation performance.

\subsection{DeCLIP}
\label{declip}
The previous section presents a method for obtaining the decoupled ``context'' and ``content'' features from CLIP. In this section, we elaborate on how the decoupled features $\mathbf{X}_{\text{content}}$ and $\mathbf{X}_{\text{context}}$ learn from their respective teacher models to enhance CLIP's performance on open-vocabulary dense prediction tasks.
\par \mypara{Content feature distillation.} As shown in Figure \ref{fig5}, the first teacher model in DeCLIP is itself, which is known as self-distillation \cite{silc,wu2023clipself,ZeroSeg,pacl}. we employ an image patching method to align the region representations of the student model’s feature map with the corresponding image crop representations (i.e., [CLS] token) of the teacher model. 

\par Specifically, the input image $\mathbf{I}$ is first divided into $k$ sub-regions. Subsequently, these sub-regions are cropped from the original image, resulting in a set of sub-images $S = \left\{\mathbf{I}_1^{\prime}, \mathbf{I}_2^{\prime}, \dots, \mathbf{I}_k^{\prime} \right\}$. The student model takes the image $\mathbf{I}$ as input and outputs the content feature $\mathbf{X}_{\text{content}} \in \mathbb{R}^{C \times H \times W}$ and the context feature $\mathbf{X}_{\text{context}} \in \mathbb{R}^{D \times H \times W}$, as mentioned in Eq.\eqref{eq5}. Here, $D$ represents the dimension of the CLIP visual encoder, and $C$ represents the shared dimension of the vision-language modality. Then, the student model uses RoI Align \cite{maskrcnn} to pool region features from $\mathbf{X}_{\text{content}}$ based on the cropping coordinates of $S$, resulting in a region feature set $F_s = \left\{\bm{f}_1^s, \bm{f}_2^s, \dots, \bm{f}_k^s \right\}$, where $\bm{f}_i^s \in \mathbb{R}^{1 \times C}$.
\par Meanwhile, the teacher model takes the sub-image set $S$ as input and outputs a series of [CLS] tokens corresponding to the cropped sub-images, resulting in [CLS] token set $F_t = \left\{\bm{f}_1^t, \bm{f}_2^t, \dots, \bm{f}_k^t \right\}$, where $\bm{f}_i^t \in \mathbb{R}^{1 \times C}$. Finally, we use a cosine similarity loss to align the [CLS] tokens from $F_t$ with the region features from $F_s$ as follows: 
\begin{equation}
\mathcal{L}_{\mathrm{content}}=\frac{1}{k} \sum_{i=1}^k\left(1-\frac{\bm{f}_i^t \cdot \bm{f}_i^s}{\left\|\bm{f}_i^t\right\| \cdot\left\|\bm{f}_i^s\right\|}\right).
\end{equation}
\par The intuition behind this distillation branch is that, for objects within an image, classifying them using image crops (i.e., [CLS] token) achieves higher accuracy than using region features \cite{wu2023clipself}. This is because CLIP is pre-trained on image-text pairs using contrastive learning, as mentioned in Sec.\ref{preliminary}. 
Therefore, the distillation learning of $\mathbf{X}_{\text{content}}$ enhances the discriminability of CLIP's region features, \textit{i.e.}, $F_s = \left\{\bm{f}_1^s, \bm{f}_2^s, \dots, \bm{f}_k^s \right\}$, by mimicking the [CLS] tokens obtained from the image crops, \textit{i.e.}, $F_t = \left\{\bm{f}_1^t, \bm{f}_2^t, \dots, \bm{f}_k^t \right\}$. 
However, as previously discussed in Sec.\ref{observation}, the region-level fine-tuning remains limited in image segmentation that requires pixel-wise scene understanding. 
\par \mypara{Context feature distillation.} As discussed in Sec.\ref{observation}, VFMs do not exhibit CLIP’s “proxy” token issue and better correlate semantically related image tokens, which may be conducive to the fine-grained local perception. Therefore, we distilled these correlations into CLIP’s $\mathbf{X}_{\text{context}}$ features.
\par As illustrated in Figure \ref{fig5}, the same image $\mathbf{I}$ is input into the VFM to obtain its dense feature representations $\mathbf{X}_{\text{dense}}^{\text{VFM}} \in \mathbb{R}^{D \times HW}$. To ensure consistency in the number of image tokens after patch embedding, different input resolutions are typically used for the VFM and the student CLIP. To transfer VFM's correlations between image tokens to CLIP, an intermediary is required to represent the correlation volume between two image tokens. Cosine similarity is used in our method, specifically as follows:
\begin{equation}
r_{i j}=\frac{\bm{x}_i \cdot \bm{x}_j}{\left\|\bm{x}_i\right\| \cdot\left\|\bm{x}_j\right\|}.
\end{equation}
\par Here, $\bm{x}_i \in \mathbb{R}^{1 \times D}$ and $\bm{x}_j \in \mathbb{R}^{1 \times D}$ represent the $i$-th and $j$-th image patch tokens. $r_{i j}$ denotes the correlation volume between patch tokens $\bm{x}_i$ and $\bm{x}_j$. We use the L2 loss to align the discrepancy in the correlation volume between the image tokens of $\mathbf{X}_{\text{dense}}^{\text{VFM}}$ and $\mathbf{X}_{\text{context}}$, specifically as follows:
\begin{equation}
\mathcal{L}_{\mathrm{context}}=\frac{1}{H W} \sum_{i=1}^H \sum_{j=1}^W\left\|r_{i j}^{\text{VFM}}-r_{i j}^{\text{CLIP}}\right\|_2,
\end{equation}
where $r_{i j}^{\text{VFM}}$ and $r_{i j}^{\text{CLIP}}$ denote the correlation volume between $\bm{x}_i$ and $\bm{x}_j$ for VFM and CLIP, respectively. Finally, the entire distillation learning process of DeCLIP can be expressed as follows:
% \begin{equation}
% \mathcal{L}_{\mathrm{total}}=\bm{\lambda}_1 \mathcal{L}_{\mathrm{content}}+\bm{\lambda}_2\mathcal{L}_{\mathrm{context}}.
\begin{equation}
\mathcal{L}_{\mathrm{total}}=\mathcal{L}_{\mathrm{content}}+\lambda\mathcal{L}_{\mathrm{context}},
\end{equation}
where $\lambda$\footnote{The sensitivity analysis is in the appendix.} represents the loss scaling hyperparameter.

%% file: 3_experiment.tex
\section{Experiments}
\subsection{Datasets and Evaluation}
We conducted extensive evaluations across multiple open-vocabulary dense prediction benchmarks, encompassing object detection, semantic segmentation, and segmentation based on VLM features. Due to space limitations, detailed descriptions of the datasets, evaluation metrics, and implementation specifics are provided in the Appendix.
\begin{figure*}[tbp]
    \centering
    \begin{minipage}{0.3\textwidth}
        \centering
        \includegraphics[width=\linewidth]{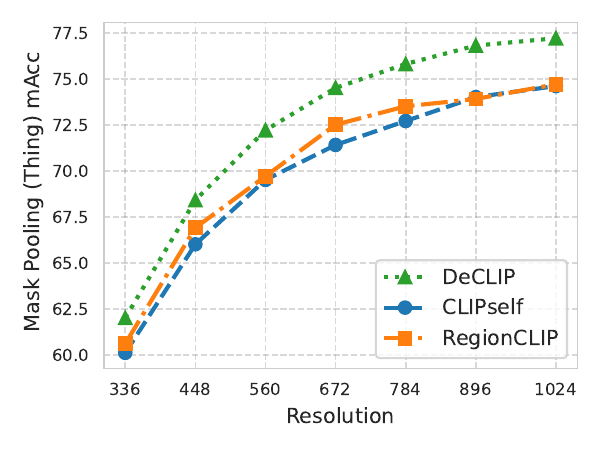}
    \end{minipage}
    \hfill
    \begin{minipage}{0.3\textwidth}
        \centering
        \includegraphics[width=\linewidth]{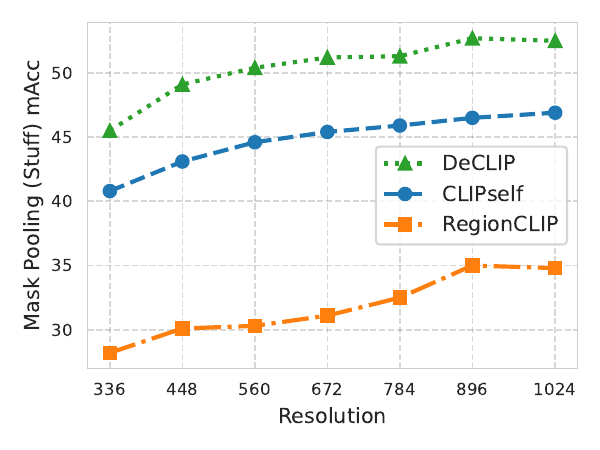}
    \end{minipage}
    \hfill
    \begin{minipage}{0.3\textwidth}
        \centering
        \includegraphics[width=\linewidth]{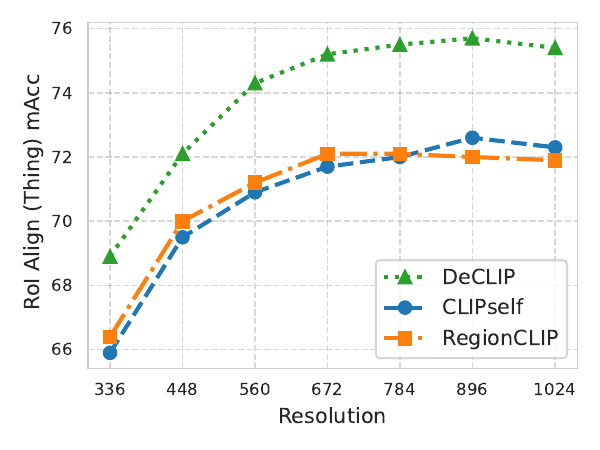}
    \end{minipage}

    \caption{Comparisons between DeCLIP and existing methods in terms of open-vocabulary region classification ability at different resolutions on the COCO panoptic dataset.}
    \label{fig6}
\end{figure*}

\subsection{Benchmark Results}

\begin{table*}[htbp]
\caption{Comparison with state-of-the-art open-vocabulary object detection methods. Caption supervision indicates that the method learns from extra image-text pairs, while CLIP supervision refers to transferring knowledge from CLIP. $^\dagger$: DETR-based detectors \cite{detr}.}
\label{tab1}
  \centering
    \renewcommand{\arraystretch}{0.86}
  \begin{subtable}[t]{0.45\textwidth}
  \centering
    \caption{OV-COCO benchmark}
      \begin{adjustbox}{width=\linewidth,center,valign=t}
      \begin{tabular}{l|l|l|c}
        \toprule
        Method & Supervision & Backbone & $\text{AP}_{50}^{\text{Novel}}$ \\
        \midrule
        ViLD \cite{vild} & CLIP & RN50  & 27.6 \\
        Detic \cite{detic} & Caption   & RN50 & 27.8 \\
        OV-DETR$^\dagger$ \cite{ovdetr} & CLIP     & RN50     & 29.4 \\
        BARON-KD \cite{wu2023aligning} & CLIP   & RN50     & 34.0 \\
        SAS-Det \cite{sasdet} & CLIP    & RN50 & 37.4 \\
        OV-DQUO$^\dagger$ \cite{ovdquo} & CLIP     & RN50       & 39.2 \\
        RegionCLIP \cite{regionclip} & Captions  & RN50x4       & 39.3 \\
        CORA$^\dagger$ \cite{wu2023cora} & CLIP   & RN50x4    & 41.7 \\
        OV-DQUO$^\dagger$ \cite{ovdquo} & CLIP     & RN50x4       & 45.6 \\
        \midrule
        RO-ViT \cite{kim2023region} &  CLIP   & ViT-L/16     & 33.0 \\
        CFM-ViT \cite{CFM} & CLIP   & ViT-L/16     & 34.1 \\
        CLIPSelf \cite{wu2023clipself} & CLIP  & ViT-B/16  & 37.6 \\
        CLIPSelf \cite{wu2023clipself} & CLIP  & ViT-L/14  & 44.3 \\
        \midrule
        \rowcolor[HTML]{f0f0f0}F-ViT \cite{wu2023clipself}+DeCLIP & CLIP     & ViT-B/16       &41.1   \textcolor[HTML]{2ECC71}{(+3.5)}\\
        \rowcolor[HTML]{EFEFEF}F-ViT \cite{wu2023clipself}+DeCLIP & CLIP     & ViT-L/14       &46.2  \textcolor[HTML]{2ECC71}{(+1.9)}  \\
       \rowcolor[HTML]{EFEFEF}OV-DQUO+DeCLIP$^\dagger$ & CLIP     & ViT-B/16       & \textbf{46.1} \textcolor[HTML]{808080}{(+6.9)}  \\
        \rowcolor[HTML]{EFEFEF}OV-DQUO+DeCLIP$^\dagger$ & CLIP     & ViT-L/14       & \textbf{48.3} \textcolor[HTML]{808080}{(+2.7)} \\
        \bottomrule
      \end{tabular}
    \end{adjustbox}
  \end{subtable}
  \hspace{10mm}
  \begin{subtable}[t]{0.452\textwidth}
  \centering
    \caption{OV-LVIS benchmark}
    \begin{adjustbox}{width=\linewidth,center,valign=t}
      \begin{tabular}{l|l|l|c}
        \toprule
        Method & Supervision & Backbone & $\text{mAP}_{r}$ \\
        \midrule
        ViLD \cite{vild} &CLIP & RN50 & 16.3 \\
        OV-DETR$^\dagger$ \cite{ovdetr} &CLIP & RN50 & 17.4 \\
        BARON-KD \cite{wu2023aligning} &CLIP & RN50 & 22.6 \\
        RegionCLIP \cite{regionclip} &Caption & RN50x4 & 22.0 \\
        OV-SAM \cite{ovsam} &CLIP & RN50x16 & 24.0 \\
        CORA$^{+}$$^\dagger$ \cite{wu2023cora} &Caption & RN50x4 & 28.1 \\
        F-VLM \cite{fvlm} &CLIP & RN50x64 & 32.8 \\
        \midrule
        CLIPSelf \cite{wu2023clipself} &CLIP & ViT-B/16 & 25.3 \\
        OV-DQUO$^\dagger$ \cite{ovdquo} &CLIP & ViT-B/16 & 29.7 \\
        Detic \cite{detic} & Caption   & Swin-B & 33.8 \\
        RO-ViT \cite{kim2023region} &CLIP & ViT-H/16 & 34.1 \\
        CLIPSelf \cite{wu2023clipself} &CLIP & ViT-L/14 & 34.9 \\
        OV-DQUO$^\dagger$ \cite{ovdquo} &CLIP & ViT-L/14 & 39.3 \\
        \midrule
        \rowcolor[HTML]{EFEFEF}F-ViT \cite{wu2023clipself}+DeCLIP & CLIP     & ViT-B/16       & 26.8 \textcolor[HTML]{2ECC71}{(+1.5)} \\
        \rowcolor[HTML]{EFEFEF}F-ViT \cite{wu2023clipself}+DeCLIP & CLIP     & ViT-L/14       & 37.2 \textcolor[HTML]{2ECC71}{(+2.3)}  \\
        \rowcolor[HTML]{EFEFEF}OV-DQUO+DeCLIP$^\dagger$ & CLIP     & ViT-B/16       &\textbf{31.0} \textcolor[HTML]{2ECC71}{(+1.3)}  \\
        \rowcolor[HTML]{EFEFEF}OV-DQUO+DeCLIP$^\dagger$ & CLIP     & ViT-L/14       & \textbf{41.5} \textcolor[HTML]{2ECC71}{(+2.2)}  \\
        \bottomrule
      \end{tabular}
    \end{adjustbox}
  \end{subtable}
\label{tab2}
\end{table*}

\begin{table}[tbp]
\centering
\renewcommand{\arraystretch}{0.88}
\caption{Transfer evaluation of the LVIS-trained detector on COCO and Objects365 datasets. }
\begin{adjustbox}{width=\linewidth,center,valign=t}
\begin{tabular}{l|ccc|ccc}
\toprule 
\multirow{2.5}{*}{Method} & \multicolumn{3}{c|}{COCO} & \multicolumn{3}{c}{Objects365 \cite{object365}} \\
\cmidrule(lr){2-4} \cmidrule(lr){5-7} & AP & AP$_{50}$ & AP$_{75}$ & AP & AP$_{50}$ & AP$_{75}$ \\
\midrule
Supervised Baseline \cite{vild} & 46.5 & 67.6 & 50.9 & 25.6 &38.6  & 28.0 \\
\midrule
ViLD \cite{vild} & 36.6 &55.6  & 39.6 & 11.8 &18.0  & 12.6 \\
DetPro \cite{du2022learning} & 34.9 &53.8  & 37.4 & 12.1 &18.8  & 12.9 \\
BARON \cite{wu2023aligning} & 36.2 &55.7  & 39.1 & 13.6 & 21.0 & 14.5 \\
F-VLM \cite{fvlm} & 37.9 &61.6  & 41.2 & 16.2 & 27.4 & 17.5 \\
CoDet \cite{codet} & 39.1 &57.0  & 42.3 & 14.2 &20.5  & 15.3 \\
RO-ViT \cite{rovit} & - &-  & - & 17.7&27.4  & 19.1 \\
CLIPSelf \cite{wu2023clipself} & 40.5 &63.8  & 44.3 & 19.5 &31.3  & 20.7 \\
\midrule
 \rowcolor[HTML]{EFEFEF}DeCLIP &\textbf{41.0}  &\textbf{64.6}  &\textbf{44.8}  &\textbf{20.0} &\textbf{32.2}  &\textbf{21.2}  \\
\bottomrule
\end{tabular}
\end{adjustbox}
\label{tab3}
\end{table}

\mypara{Open-Vocabulary Detection.} Table \ref{tab1} presents DeCLIP’s performance on OV-COCO and OV-LVIS benchmarks. On OV-COCO, DeCLIP improves the F-ViT \cite{wu2023clipself} baseline by 3.5 and 1.9 mAP, and the OV-DQUO \cite{ovdquo} baseline by 6.9 and 2.7 mAP on novel classes. On OV-LVIS, it achieves gains of 1.5 and 2.3 mAP with F-ViT, as well as 1.3 and 2.2 mAP with OV-DQUO on rare classes. Cross-dataset evaluations of F-ViT+DeCLIP trained on OV-LVIS (Table \ref{tab3}) further confirm DeCLIP’s superiority over existing methods.

\mypara{Open-Vocabulary Semantic Segmentation.} Table \ref{tab4} displays the performance of the CAT-Seg \cite{catseg} model using DeCLIP as the backbone across various open-vocabulary semantic segmentation benchmarks. The results show that DeCLIP significantly enhances segmentation performance on all datasets. Notably, even with the ViT-B/16 version of DeCLIP, CAT-Seg nearly surpasses all existing SOTA methods that utilize substantially larger encoders like ConvNeXt-L. When employing the ViT-L/14 version of DeCLIP, the model achieves new SOTA results in open-vocabulary semantic segmentation tasks.
\begin{table*}[htbp]
    \centering
    \renewcommand{\arraystretch}{0.92}
    \caption{Results on open-vocabulary semantic segmentation. $^\dagger$ indicates results re-experimented by CAT-Seg \cite{catseg}.}
    \begin{adjustbox}{width=.92\textwidth}
    \begin{tabular}{l|ll|ccccccc}
    \toprule
    Method & Backbone & Training Set & ADE847 & Context459 & ADE150 & Context59 & VOC20 & VOC21\\
    \midrule
    ZegFormer$^\dagger$ \cite{ZegFormer} &ViT-B/16  & COCO-Stuff   &5.6   &10.4   & 18.0   &45.5   &89.5   &65.5   \\
    ZSseg \cite{ZSseg}    &ViT-B/16  & COCO-Stuff   &7.0   &-   &20.5   &47.7   & 88.4   &-   \\
    OVSeg \cite{OVSeg}     &ViT-L/14  & COCO-Stuff   &9.0   &12.4   & 29.6   & 55.7   &  94.5   &-  \\
    SAN \cite{SAN}     &ViT-L/14  & COCO-Stuff   &13.7   &17.1   &33.3   &60.2   &95.5   &-   \\
    ODISE \cite{ODISE}     &ViT-L/14  & COCO-Panoptic &11.1   &14.5   & 29.9   & 57.3   &-   &84.6   \\
    MAFT \cite{MAFT}& ConvNeXt-L  &COCO-Stuff  &13.1 &17.0  &34.4  & 57.5   &93.0   &-  \\
    FC-CLIP \cite{fcclip}  & ConvNeXt-L & COCO-Panoptic & 14.8   &18.2  &34.1   & 58.4   & 95.4   & 81.8   \\
    FrozenSeg \cite{frozenseg}& ConvNeXt-L & COCO-Panoptic &14.8 &19.7 &34.4 &- &-&82.5\\
    CAT-Seg \cite{catseg} &ViT-B/16  &COCO-Stuff  &12.0  &19.0  &31.8  &57.5   &94.6   &77.3   \\
    CAT-Seg \cite{catseg}  &ViT-L/14  &COCO-Stuff  &16.0  &23.8  &37.9  &63.3   &97.0   &82.5   \\    \midrule
    \rowcolor[HTML]{EFEFEF}  
    CAT-Seg+DeCLIP & ViT-B/16  & COCO-Stuff  & 15.3 \textcolor[HTML]{2ECC71}{(+3.3)}  & 21.4 \textcolor[HTML]{2ECC71}{(+2.4)} & 36.3 \textcolor[HTML]{2ECC71}{(+4.5)}  & 60.6 \textcolor[HTML]{2ECC71}{(+3.1)} & 96.6 \textcolor[HTML]{2ECC71}{(+2.0)} & 81.3 \textcolor[HTML]{2ECC71}{(+4.0)} \\
   \rowcolor[HTML]{EFEFEF}  CAT-Seg+DeCLIP & ViT-L/14  & COCO-Stuff  & \textbf{17.6} \textcolor[HTML]{2ECC71}{(+1.6)}  & \textbf{25.9} \textcolor[HTML]{2ECC71}{(+2.1)} & \textbf{40.7} \textcolor[HTML]{2ECC71}{(+2.8)}  & \textbf{63.9} \textcolor[HTML]{2ECC71}{(+0.6)} & \textbf{97.7} \textcolor[HTML]{2ECC71}{(+0.7)}  & \textbf{83.9} \textcolor[HTML]{2ECC71}{(+1.4)} \\
    \bottomrule
    \end{tabular}
    \end{adjustbox}
\label{tab4}
\end{table*}

\mypara{Open-Vocabulary Semantic Segmentation Based on VLM Features.} Following existing methods \cite{CLIPtrase,sclip,clearclip}, in this experiment, we assign each pixel in the feature map the category with which it has the highest cosine similarity. The low-resolution prediction result is up-sampled to the original resolution to obtain the final segmentation map. As shown in Table \ref{tab5}, DeCLIP outperforms all existing methods in terms of average mIoU across eight benchmarks, highlighting the effectiveness of our approach in improving the discriminability and spatial consistency of VLM features. 

% We assigned each pixel to the category with the highest cosine similarity and upsampled the prediction to the original resolution. During inference, images were resized to a shorter side of 448 pixels, using a sliding window of 336×336 with a stride of 112×112, without post-processing. As shown in Table \ref{tab5}, DeCLIP outperforms all existing methods in average mIoU across eight benchmarks, demonstrating its effectiveness in enhancing the discriminability and spatial consistency of VLM features.

\begin{table*}[htbp]
    \renewcommand{\arraystretch}{0.88}
    \centering
    \caption{Results on open-vocabulary semantic segmentation based on VLM features.}
    \begin{adjustbox}{width=.92\textwidth}
    \begin{tabular}{lccc|ccccccc|c}
        \toprule
        \multirow{2.5}{*}{Method} & \multicolumn{3}{c}{With a background category} & \multicolumn{5}{c}{Without background category} & \multirow{2.5}{*}{Avg} \\
        \cmidrule(lr){2-4} \cmidrule(lr){5-9}
        & VOC21  & Context60 & COCO-Obj & VOC20 & CityScape & Context59 & ADE & COCO-Stf &  \\
        \midrule
        CLIP \cite{clip} & 18.8 & 9.9  & 8.1  & 49.4 & 6.5  & 11.1 & 3.1  & 5.7  & 14.1 \\
        \midrule
        MaskCLIP \cite{maskclip} & 43.4 & 23.2 & 20.6 & 74.9 & 24.9 & 26.4 & 11.9 & 16.7 & 30.3 \\
        GroupViT \cite{groupvit} & 52.3 & 18.7 & 27.5 & 79.7 & 18.5 & 23.4 & 10.4 & 15.3 & 30.7 \\
        ReCo \cite{reco} & 25.1 & 19.9 & 15.7 & 57.7 & 21.6 & 22.3 & 11.2 & 14.8 & 23.5 \\
        TCL \cite{tcl}& 51.2 & 24.3 & 30.4 & 77.5 & 23.5 & 30.3 & 14.9 & 19.6 & 33.9 \\
        OVSeg \cite{OVSeg}& 53.8 & 20.4 & 25.1 & -    & -    & -    & 5.6  & -    & -  \\
        SCLIP \cite{sclip}& 59.1 & 30.4 & 30.5 & 80.4 & \underline{32.2} & 34.2 & 16.1 & 22.4 & 38.2 \\
        ClearCLIP \cite{clearclip} &51.8  &\underline{32.6}  &33.0 &\underline{80.9}  &30.0  &\underline{35.9}  &16.7  &23.9  &38.1  \\
        CLIP-DINOiser \cite{clipdino}&\textbf{62.1}  &32.4  &\underline{34.8} &\underline{80.9}  &31.7  &\underline{35.9}  &\underline{20.0}  &\underline{24.6}  &\underline{40.3} \\ \midrule
        \rowcolor[HTML]{EFEFEF} 
        DeCLIP (ours) & \underline{59.7}  &\textbf{35.3}  &\textbf{36.4}  & \textbf{85.0} &\textbf{32.8}  &\textbf{39.2} &\textbf{21.9}  &\textbf{25.3} & \textbf{41.9} \\
        \bottomrule
    \end{tabular}
    \end{adjustbox}
\label{tab5}
\end{table*}

\mypara{Open-Vocabulary Region Classification.} We assess the region classification performance of DeCLIP, RegionCLIP \cite{regionclip}, and CLIPSelf \cite{wu2023clipself} at various resolutions on the COCO-Panoptic validation set. Using RoI Align \cite{maskrcnn} and Mask Pooling, we extract local features from the feature maps based on annotated bounding boxes and masks, assigning categories based on maximum cosine similarity. As illustrated in Figure \ref{fig6}, the Top-1 mean accuracy (mAcc) results demonstrate that DeCLIP consistently surpasses existing methods in region recognition across all resolutions.
\begin{figure}[htbp]
  \centering
  \includegraphics[width=.95\linewidth]{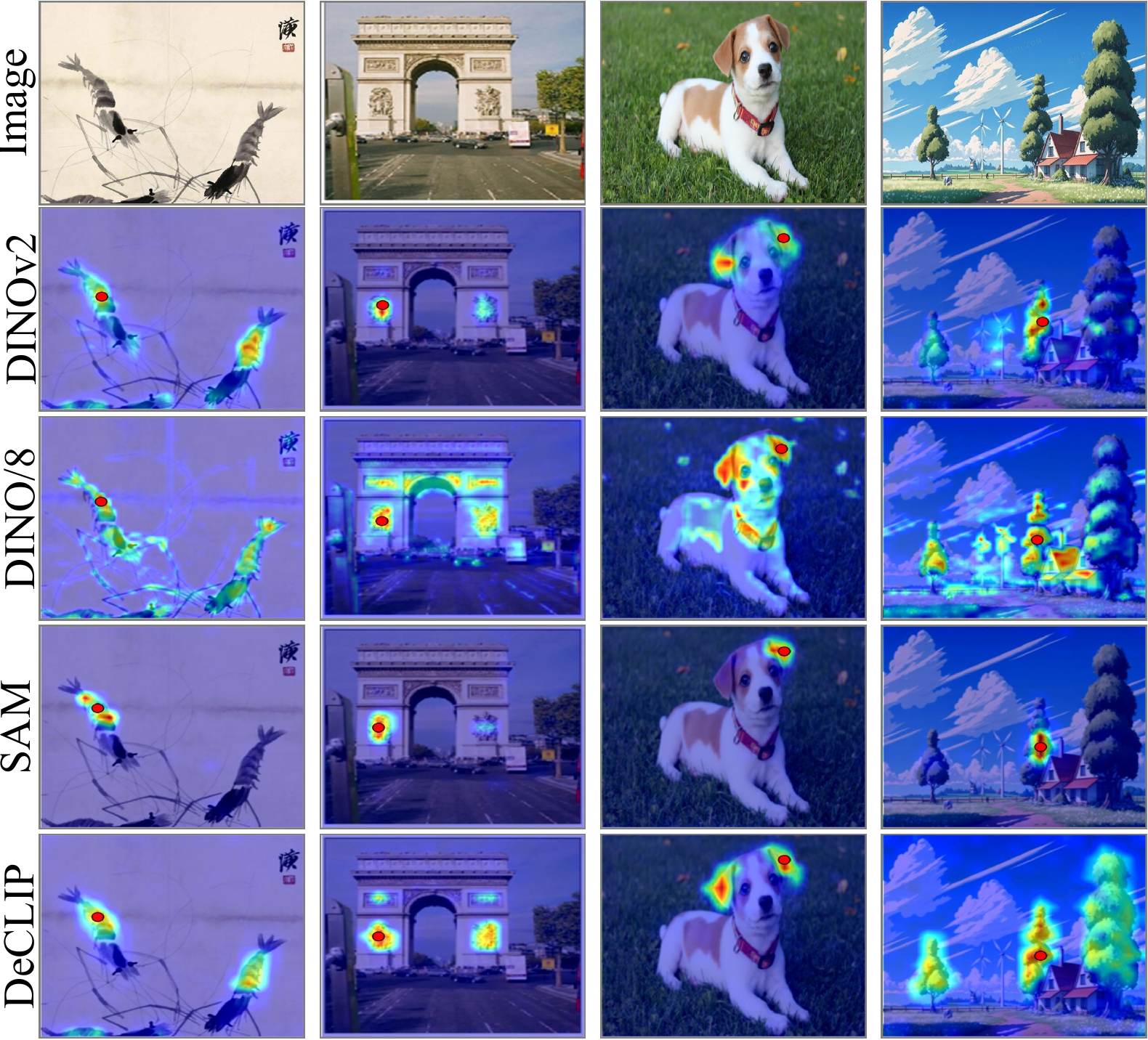}
  \caption{Qualitative comparisons of attention maps between VFMs and DeCLIP. The anchor image token is marked in red. }
  \label{fig8}
\end{figure}
\begin{table}[thbp]
    \centering
    \caption{Ablation studies on the impact of different VFMs on open-vocabulary region classification and segmentation.}
    \begin{adjustbox}{width=\linewidth,center,valign=t}
    \begin{tabular}{ll|ccccc}
        \toprule
          \multirow{2.5}{*}{VFMs} &\multirow{2.5}{*}{Arch} &  \multicolumn{2}{c}{Region Classification (mAcc)} & \multicolumn{3}{c}{Semantic Segmentation (mIoU)} \\
        \cmidrule(lr){3-4}\cmidrule(lr){5-7}
        &  & COCO (Thing) & COCO (Stuff) & Context59 & COCO-Stf & ADE \\
        \midrule
        DINO \cite{dino} & ViT-B/8 &68.4 &49.4 &37.3 &23.2 &19.5 \\
        DINO \cite{dino} & ViT-B/16 &67.6 &47.4 &38.1 &23.7 &20.4 \\
        SAM \cite{sam} & ViT-B/16 &75.0 &51.8 &35.3 &22.0 &18.5 \\
        SAM \cite{sam} & ViT-L/16 &76.8 &52.6 &37.7 &23.0 &20.0 \\
        DINOv2 \cite{dinov2} & ViT-B/14 & 77.2  &52.5  &\textbf{39.2} &\textbf{25.3} &\textbf{21.9} \\
        DINOv2 \cite{dinov2} & ViT-L/14 &\textbf{77.6} &\textbf{53.1} &38.0 &24.1 &21.3 \\
        \bottomrule
    \end{tabular}
    \end{adjustbox}
\label{tab7}
\end{table}

\subsection{Ablation Study}
\mypara{The impact of VFMs.} We analyzed the impact of various VFM configurations on DeCLIP performance. As shown in Table \ref{tab7}, DeCLIP distilled from DINO \cite{dino} performs moderately in segmentation but trails SAM \cite{sam,sam2} and DINOv2 \cite{dinov2} in region classification. DeCLIP distilled from SAM excels in region classification but shows lower segmentation performance compared to DINO. DINOv2 achieves balance in both region classification and segmentation.

\mypara{Qualitative results.} Figure \ref{fig8} presents the visual comparison of attention maps between DINO, SAM, DINOv2, and DeCLIP. Experimental results show that DeCLIP effectively focuses on regions spatially or semantically associated with the anchor image token. Moreover, this experiment reveals why DeCLIP distilled from DINOv2 works best: SAM lacks semantic association ability, while DINO focus indiscriminately on all primary objects in the image. 

%% file: 5_conclusion.tex
\section{Conclusion}
This paper analyzes the limitations of CLIP in dense prediction tasks from the perspective of its attention map. We observed that CLIP's [CLS] token negatively affects the attention map of image tokens. To address this issue, we proposed DeCLIP, a decoupled feature enhancement strategy. Extensive experiment results on open-vocabulary dense prediction benchmarks demonstrate that DeCLIP outperforms state-of-the-art methods, achieving excellent performance across all evaluated task domains.

%% file: 6_suppl.tex
% \appendix
% \setcounter{page}{1}
\maketitlesupplementary
\section*{Overview}
This material provides supplementary details to the main paper, including the following sections:
\vspace{0.3cm}
\begin{itemize}[itemsep=1pt, parsep=5pt]
\item (\ref{motivation}) \textbf{Details of Proxy Token Phenomenon}
\item (\ref{experiment}) \textbf{Additional Experiments}
\begin{itemize}
\item (\ref{ablation}) Ablation Studies
\item (\ref{app_sanity}) sanity Checks
\item (\ref{benchmark}) Further Details on Benchmark Results
\end{itemize}
\item (\ref{qualitative_analysis}) \textbf{Additional Qualitative Analysis}
\begin{itemize}
\item (\ref{corelation}) Analyses of Feature Correlations
\item (\ref{semantic_result}) Comparison of Semantic Segmentation Results
\item (\ref{attention_map}) Comparison of Attention Maps
\end{itemize}
\item (\ref{exp_settings}) \textbf{Details of Experimental Settings}
\begin{itemize}
\item (\ref{dataset}) Datasets and Evaluation Protocols
\item (\ref{detail}) Implementation Details  
\end{itemize}
\item (\ref{related_work}) \textbf{Related Work}
\begin{itemize}
\item (\ref{dense_prediction}) Open-Vocabulary Dense Prediction
\item (\ref{transfer}) Transferring VLMs to Dense Prediction Tasks
\item (\ref{vfm}) Vision Foundation Models
\end{itemize}
\end{itemize}
\section{Details of Proxy Token Phenomenon}
\label{motivation}
This section primarily supplements the details of the proxy token phenomenon observed in CLIP, offering deeper insights into the rationale behind our proposed DeCLIP.

\mypara{Observation.} As stated in the main paper, ViT-based~\cite{ViT} CLIP utilizes the [CLS] token to represent the overall features of an image and performs image-text contrastive learning accordingly. Therefore, it is commonly believed that the [CLS] token comprehensively attends to all image tokens during the forward pass to obtain a ``global view", thereby enhancing the image classification process.

\par
\begin{figure*}[htbp]
\centering
\includegraphics[width=\linewidth]{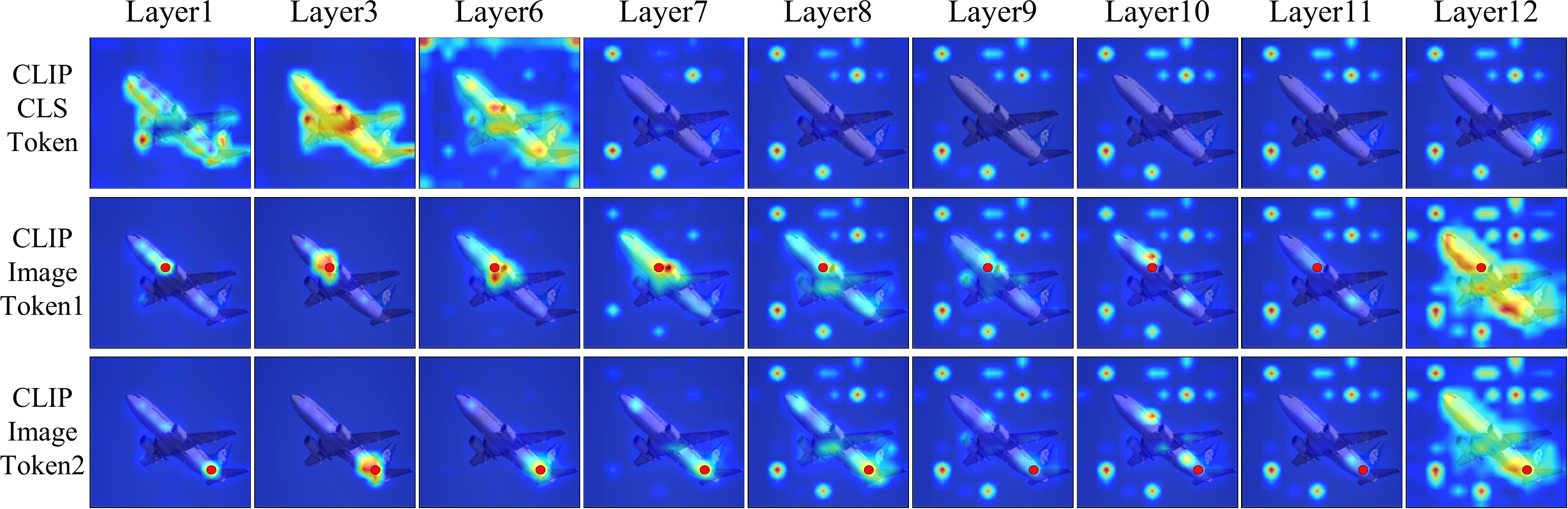}
\caption{\textbf{Visualization of the ``proxy" token phenomenon in the attention maps of the CLIP visual encoder.} Specifically, the input image resolution is 224*224. We extract the attention weights from each attention block of CLIP and average them across the multi-head dimension (after Softmax), yielding attention maps $\mathbf{M} \in \mathbb{R}^{197 \times 197}$. $\mathbf{M}\text{[0,~1:]}\in \mathbb{R}^{1 \times 196}$ represents the attention map from the [CLS] token to other image tokens (first row). $\mathbf{M}\text{[1:197,~1:197]}\in \mathbb{R}^{196 \times 196}$ represents the attention map between each image token and all image tokens. We randomly select specific image tokens' attention map (the second and third rows, indicated by the red dots) for visualization, each with dimensions of 1*196. We reshape them to 1*14*14 and apply bilinear upsampling to 1*224*224 for better visualization.}
\label{app_fig1}
\end{figure*}

 Unexpectedly, the [CLS] token ceased to focus on the primary object in the image starting from the 7th layer and instead redirected its attention to several image tokens in the background as shown in the first row of Figure~\ref{app_fig1}. These specific image tokens continued to receive significant attention from the [CLS] token in the following encoding layers. 
\par A similar pattern was observed in the attention maps of CLIP's image tokens. As shown in the second row of Figure~\ref{app_fig1}, we first randomly selected an image token located on the primary object in the image as the anchor image token, and then visualized its attention maps across different encoder layers. The experimental results show that the attention of the anchor image token in layers 1-6 is primarily distributed over the object it belongs to. However, after the 7th layer, which is when the [CLS] token shifted its attention to several specific image tokens in the background, the anchor image token also began to focus on these specific image tokens.
\par Moreover, as illustrated in the third row of Figure~\ref{app_fig1}, when the position of the anchor image token is shifted, the new anchor image token continues to exhibit high attention towards these specific tokens. This demonstrates that this phenomenon is not limited to a particular image token but is instead widespread across the image tokens in CLIP.

\mypara{Analysis.} One possible explanation for this phenomenon could be the redundancy present in image data. Images inherently carry a higher information load than text, encompassing substantial background details that are unrelated to image classification tasks. These specific background tokens may serve as ``proxies'' for the [CLS] token. This suggests that these tokens aggregate essential information from other image tokens, enabling the [CLS] token to form an approximate ``global view'' by summarizing content from them, thereby facilitating image classification. This perspective is also supported by recent studies~\cite{CLIPtrase,register}.
\par In over a decade of CNN~\cite{resnet,convnext} development, no studies have reported similar phenomena. Therefore, we speculate that the second reason for this phenomenon may stem from the ViT architecture~\cite{ViT}. The classic ResNet~\cite{resnet} architecture consists of four stages, in which the feature resolution is halved and the number of channels is doubled at each stage. This is a process of learning sparse features, where redundant image details are progressively discarded, and feature semantics are continually enhanced. However, CLIP with a ViT architecture lacks this process. After patch embedding, the size and the number of channels in the feature map remain unchanged. As a result, the model spontaneously generates ``proxy" tokens to mimic the process of learning sparse features, akin to CNN. 

\mypara{Effects.} As discussed above, the proxy token phenomenon allows ViT CLIP to learn sparse features, which facilitate the extraction of key information from images, enhance image-text contrastive learning and reduce the optimization burden.

\par However, this phenomenon causes the image tokens in CLIP to indiscriminately focus on the proxy tokens in the background, rather than on the regions that are spatially or semantically related to them. Consequently, this leads to CLIP's dense features to lack local discriminability and spatial consistency, affecting its performance in open-vocabulary dense prediction tasks. 

\begin{table}[tbp]
    \centering
    \caption{Ablation study on types of $\mathbf{X}_\text{Context}$.}
    \begin{adjustbox}{width=\linewidth,center,valign=t}
    \begin{tabular}{c|cccc}
        \toprule
        \multirow{2.5}{*}{$\mathbf{X}_\text{Context}$} & \multicolumn{2}{c}{Region Classification (mAcc)} & \multicolumn{2}{c}{Semantic Segmentation (mIoU)} \\
        \cmidrule(lr){2-3} \cmidrule(lr){4-5}
         & COCO (Thing) & COCO (Stuff) &  PASCAL Context59 & ADE \\
        \midrule
        $\mathbf{Q}$ & 77.2  &52.5  &38.7 &21.8 \\
        $\mathbf{K}$&76.5  &51.0  &\textbf{39.4} &21.6 \\
        \rowcolor[HTML]{EFEFEF} $\mathbf{Q}+\mathbf{K}$ &\textbf{77.3}  &\textbf{53.8}  &39.2 &\textbf{21.9} \\
        \bottomrule
    \end{tabular}
    \end{adjustbox}
\label{app_tab1}
\end{table}
\section{Additional Experiments}
\label{experiment}
\subsection{Ablation Studies}
\label{ablation}
In this section, we conduct a thorough ablation study on DeCLIP, encompassing the examination of various $\mathbf{X}_{\text{context}}$ implementations, the variation in the number of fine-tuning layers, the impact of the hyperparameter $\lambda$ in the loss function, and the influence of the distillation baseline.
\par Except for the region classification experiment in Table~\ref{app_tab1}, which was conducted at a resolution of 1024×1024, the region classification performance in all other experiments was assessed at a resolution of 560×560. Additionally, the semantic segmentation performance of all ablation experiments was assessed at a resolution of 336×336.

\mypara{Types of Context.} Since there are various implementations of $\mathbf{X}_{\text{context}}$, including $\mathbf{Q}$, $\mathbf{K}$, and $\mathbf{Q}+\mathbf{K}$, we performed an ablation study on their performance in dense prediction tasks, including region classification (mAcc) and semantic segmentation (mIoU), as shown in Table \ref{app_tab1}. Specifically, implementing $\mathbf{X}_{\text{context}}$ based on $\mathbf{K}$ means that the last attention block of CLIP leverages $\mathbf{K}$ to compute the attention weight. Additionally, implementing $\mathbf{X}_{\text{context}}$ based on $\mathbf{Q}+\mathbf{K}$ involves first computing the attention weights of $\mathbf{Q}$ and $\mathbf{K}$ separately, and then summing them. The experimental results indicate that the performance differences among the three implementations are minimal, while the $\mathbf{Q}$ and $\mathbf{K}$ exhibits slightly better performance in dense prediction tasks.

\begin{table}[tbp]
    \centering
    \caption{Ablation study on number of fine-tuning layers.}
    \begin{adjustbox}{width=\linewidth,center,valign=t}
    \begin{tabular}{c|cccc}
        \toprule
        \multirow{2.5}{*}{\makecell{Fine-tuning \\ Layers}} & \multicolumn{2}{c}{Region Classification (mAcc)} & \multicolumn{2}{c}{Semantic Segmentation (mIoU)} \\
        \cmidrule(lr){2-3} \cmidrule(lr){4-5}
         & COCO (Thing) & COCO (Stuff) &  PASCAL Context59 & ADE \\
        \midrule
        3 &62.7  &47.0  &38.0 &21.8 \\
        6 &67.1  &47.8  &\textbf{39.0} &\textbf{22.3} \\
        9 &70.7   &50.5  &\textbf{39.0} &22.1 \\
        \rowcolor[HTML]{EFEFEF} 12 &\textbf{72.2}  &\textbf{51.3}  &38.7 &21.8 \\
        \bottomrule
    \end{tabular}
    \end{adjustbox}
\label{tab9}
\end{table}

\mypara{Number of fine-tuning layers.} We performed an ablation study to examine the relationship between the number of fine-tuning attention blocks and dense prediction performance. The experiment was conducted on the ViT-B version of CLIP, which comprises a total of 12 attention blocks. we experiment with updating the last 3, 6, 9, and 12 attention blocks. As shown in Table~\ref{tab9}, we observed that as the number of fine-tuning layers increased, the performance of region classification continuously improved, reaching its peak at 12 layers. However, the performance of semantic segmentation peaked at 6 layers, and as the number of layers increased further, the performance slightly declined. In practice, to balance the performance of both tasks, we chose to fine-tune all attention blocks in the implementation of DeCLIP.

\begin{table*}[htbp]
    \renewcommand{\arraystretch}{0.85}
    \centering
    \caption{Ablation Study on EVA-CLIP for open-vocabulary semantic segmentation}
    \begin{adjustbox}{width=.95\textwidth}
    \begin{tabular}{l|ll|ccccccc}
    \toprule
    Method & Backbone & Training Set & ADE847 & Context459 & ADE150 & Context59 & VOC20 & VOC21\\
    \midrule
    CAT-Seg+CLIP \cite{clip} &ViT-B/16  &COCO-Stuff   &12.0  &19.0  &31.8  &57.5   &94.6   &77.3   \\
    CAT-Seg+CLIP \cite{clip}  &ViT-L/14  &COCO-Stuff  &16.0  &23.8  &37.9  &63.3   &97.0   &82.5   \\    
    \midrule
    CAT-Seg+EVA-CLIP \cite{evaclip} &ViT-B/16  &COCO-Stuff  &11.9  &17.6  &30.4  & 52.3  & 94.2  &74.2  \\
    CAT-Seg+EVA-CLIP  \cite{evaclip} &ViT-L/14  &COCO-Stuff  &14.2  &21.3  &34.8  &56.2   &95.8   &80.1   \\    
     \midrule
    \rowcolor[HTML]{EFEFEF}  
    CAT-Seg+DeCLIP & ViT-B/16  & COCO-Stuff  & 15.3   & 21.4  & 36.3   & 60.6  & 96.6  & 81.3 \\
    \rowcolor[HTML]{EFEFEF}  
    CAT-Seg+DeCLIP & ViT-L/14  & COCO-Stuff  & \textbf{17.6}  & \textbf{25.9}  & \textbf{40.7}   & \textbf{63.9}  & \textbf{97.7}  & \textbf{83.9} \\
    \bottomrule
    \end{tabular}
    \end{adjustbox}
\label{app_tab2}
\end{table*}
\begin{table*}[htbp]
    \renewcommand{\arraystretch}{0.8}
    \centering
    \caption{Ablation Study on EVA-CLIP for open-vocabulary semantic segmentation based on VLM features.}
    \begin{adjustbox}{width=.95\textwidth}
    \begin{tabular}{lccc|ccccccc|c}
        \toprule
        \multirow{2.5}{*}{Method} & \multicolumn{3}{c}{With a background category} & \multicolumn{5}{c}{Without background category} & \multirow{2.5}{*}{Avg} \\
        \cmidrule(lr){2-4} \cmidrule(lr){5-9}
        & VOC21  & Context60 & COCO-Obj & VOC20 & CityScape & Context59 & ADE & COCO-Stf &  \\
        \midrule
        CLIP \cite{clip} & 18.8 & 9.9  & 8.1  & 49.4 & 6.5  & 11.1 & 3.1  & 5.7  & 14.1 \\
        EVA-CLIP  \cite{evaclip} &23.4  &12.8  &15.3   &55.9  &12.8  &13.9 &7.7  &9.7  &18.9  \\
        \midrule
        ClearCLIP \cite{clearclip} &51.8  &32.6  &33.0 &80.9  &30.0  &35.9  &16.7  &23.9  &38.1  \\
        EVA-ClearCLIP  &47.0  &29.7  &30.2 &78.3  &26.3  &29.4  &16.7  &20.4  & 34.7  \\
        \midrule
        \rowcolor[HTML]{EFEFEF} 
        DeCLIP & \textbf{59.7}  & \textbf{35.3}  & \textbf{36.4}  & \textbf{85.0} & \textbf{32.8}  & \textbf{39.2} & \textbf{21.9} & \textbf{25.3} & \textbf{41.9} \\
        \bottomrule
    \end{tabular}
    \end{adjustbox}
\label{app_tab3}
\end{table*}

\mypara{Sensitivity Analysis of $\lambda$.} In DeCLIP, we employ a hyperparameter $\lambda$ to balance the weight between $\mathcal{L}_{\mathrm{content}}$ and $\mathcal{L}_{\mathrm{context}}$. We performed an ablation study to examine the relationship between the hyperparameter $\lambda$ and dense prediction performance. The experimental results demonstrate that our method exhibits strong robustness, and the dense prediction performance of DeCLIP does not fluctuate drastically with changes in $\lambda$. Furthermore, the results indicate that $\lambda=0.25$ strikes a good balance between region classification capability and image segmentation performance. 

\mypara{Distillation Baseline.}
In our experiments, we used EVA-CLIP~\cite{evaclip} as the baseline for DeCLIP, as we found that it demonstrated improved performance after distillation, as shown in Table~\ref{app_tab11}. This can be attributed to two main factors:  (1) EVA-CLIP uses the EVA02~\cite{eva02} model for initializing the visual encoder. EVA02 was trained using Masked Image Modeling (MIM), thereby enhancing its compatibility with Vision Foundation Models (VFMs). (2) EVA-CLIP's [CLS] token exhibits superior zero-shot classification capability compared to OpenAI's model \cite{wu2023clipself}. In Sec.~\ref{app_sanity}, we conducted comprehensive sanity checks to verify whether the performance improvement of DeCLIP in dense prediction tasks is due to the use of EVA-CLIP. 
\begin{table}[tbp]
    \renewcommand{\arraystretch}{0.9}
    \centering
    \caption{Sentitivity Analysis of hyperparameter $\lambda$.}
    \begin{adjustbox}{width=\linewidth,center,valign=t}
    \begin{tabular}{c|cccc}
        \toprule
        \multirow{2.5}{*}{$\lambda$} & \multicolumn{2}{c}{Region Classification (mAcc)} & \multicolumn{2}{c}{Semantic Segmentation (mIoU)} \\
        \cmidrule(lr){2-3} \cmidrule(lr){4-5}
         & COCO (Thing) & COCO (Stuff) &  PASCAL Context59 & ADE \\
        \midrule
        0.1 &72.4  &50.6  &37.9 &21.3 \\
        0.2 &72.4  &51.0  &38.4 &21.7 \\
         \rowcolor[HTML]{EFEFEF} 0.25 &72.2  &51.3 &38.7  &21.8 \\
        0.3 &71.9  &51.4  &38.7 &21.7 \\
        \bottomrule
    \end{tabular}
    \end{adjustbox}
\label{app_tab10}
\end{table}
\begin{figure*}[tbp]
\centering
\includegraphics[width=\linewidth]{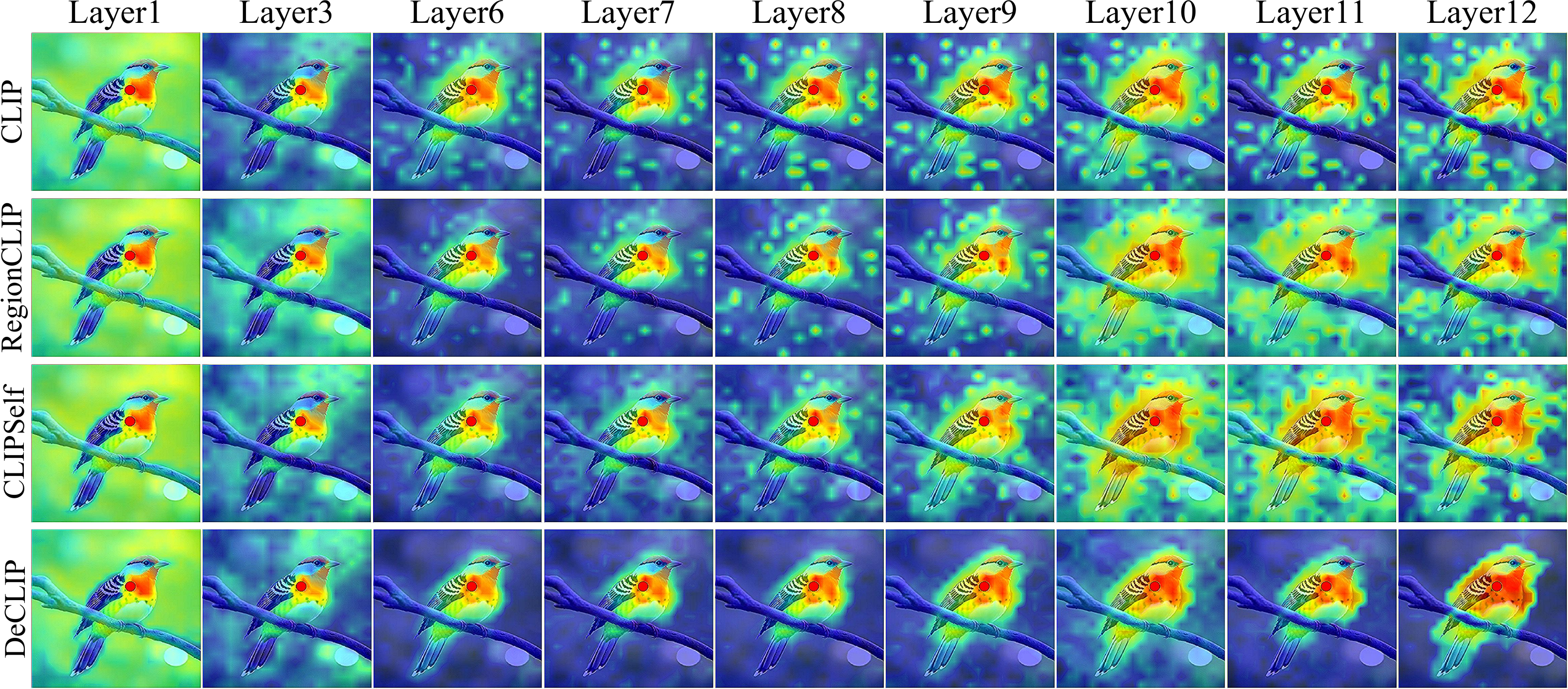}
\caption{\textbf{Qualitative comparison of feature correlations between DeCLIP and existing pre-fine-tuning approaches \cite{wu2023clipself,regionclip}.}  Specifically, the input image resolution is 336*336. We extract the output features from each attention block of CLIP, where each feature $\mathbf{F} \in \mathbb{R}^{441 \times D}$. Then, we compute the feature correlations $\mathbf{FC} \in \mathbb{R}^{441 \times 441}$ between the image tokens within $\mathbf{F}$ using cosine similarity. We randomly select a specific image token's feature correlation (indicated by the red dots) and upsample it to a resolution of 336*336 for visualization.}
\label{app_fig2}
\end{figure*}
\subsection{sanity Checks}
\label{app_sanity}
To eliminate potential biases that EVA-CLIP~\cite{evaclip} might introduce, we conducted additional sanity check experiments. 
\par Specifically, we first apply vanilla EVA-CLIP as the backbone network in the CAT-Seg~\cite{catseg} model and compare its performance with DeCLIP in the Open-Vocabulary Semantic segmentation (OVSS) task, as shown in Table~\ref{app_tab2}. Furthermore, we re-implemented ClearCLIP~\cite{clearclip} based on EVA-CLIP and named it EVA-ClearCLIP. Then, we compared the performance between EVA-CLIP, EVA-ClearCLIP, and DeCLIP in the OVSS based on VLM features task, as shown in Table~\ref{app_tab3}. We did not conduct further open-vocabulary detection experiments because the baseline detectors, OV-DQUO~\cite{ovdquo} and F-ViT~\cite{wu2023clipself}, have already used EVA-CLIP as the backbone network in their respective studies.
\begin{table}[tbp]
    \renewcommand{\arraystretch}{0.9}
    \centering
    \caption{Comparison of different distillation baselines.}
    \begin{adjustbox}{width=\linewidth,center,valign=t}
    \begin{tabular}{c|cccc}
        \toprule
        \multirow{2.5}{*}{Source} & \multicolumn{2}{c}{Region Classification (mAcc)} & \multicolumn{2}{c}{Semantic Segmentation (mIoU)} \\
        \cmidrule(lr){2-3} \cmidrule(lr){4-5}
         & COCO (Thing) & COCO (Stuff) &  PASCAL Context59 & ADE \\
        \midrule
        OpenAI &65.0&38.8  &36.2 &18.6 \\
         \rowcolor[HTML]{EFEFEF} EVA-CLIP &72.2  &51.3  &38.7 &21.8 \\
        \bottomrule
    \end{tabular}
    \end{adjustbox}
\label{app_tab11}
\end{table}

\mypara{OVSS.} As shown in Table~\ref{app_tab2}, experimental results demonstrate that directly applying EVA-CLIP to CAT-Seg performs worse than OpenAI's model. In contrast, DeCLIP significantly improves CAT-Seg's performance across all semantic segmentation benchmarks. 

\mypara{OVSS based on VLM feautures.} As shown in Table~\ref{app_tab3}, experimental results indicate that EVA-CLIP performs slightly better than CLIP in this task, while EVA-ClearCLIP underperforms in comparison to ClearCLIP. However, both EVA-CLIP and EVA-ClearCLIP fall significantly short of DeCLIP's average performance of 41.9 across the eight benchmarks.

\par Based on the results of the aforementioned experiments, we conclude that the performance improvement of DeCLIP is not attributable to the introduction of EVA-CLIP, but is instead due to the superiority of the decoupled feature enhancement strategy. 

\subsection{Further Details on Benchmark Results}
\label{benchmark}
We present detailed results for the OV-COCO, OV-LVIS, and cross-dataset benchmarks to provide a comprehensive comparison of the open-vocabulary object detection task, as shown in Tables \ref{app_tab4} and \ref{app_tab5}. 
\section{Additional Qualitative Analysis}
\label{qualitative_analysis}
This section further presents a qualitative experimental analysis of our proposed DeCLIP method in comparison to existing methods, including feature correlation analysis, semantic segmentation results, and attention map comparisons, thereby providing a more comprehensive demonstration of the superiority of DeCLIP's decoupled feature enhancement strategy.
\begin{table*}[tbp]
\caption{Detailed comparison on OV-COCO and OV-LVIS benchmarks. Caption supervision indicates that the method learns from extra image-text pairs, while CLIP supervision refers to transferring knowledge from CLIP. $^\dagger$: Detection Transformer based detectors.}
\centering
\begin{subtable}[t]{0.44\textwidth}
  \centering
    \caption{OV-COCO benchmark \cite{mscoco}} 
      \begin{adjustbox}{width=\linewidth,center,valign=t}
      \begin{tabular}{l|l|l|ccc}
        \toprule
        Method & Supervision & Backbone & $\text{AP}_{50}^{\text{Novel}}$ & \textcolor{gray}{$\text{AP}_{50}^{\text{Base}}$} & \textcolor{gray}{$\text{AP}_{50}$} \\
        \midrule
        ViLD \cite{vild} & CLIP & RN50  & 27.6 & \textcolor{gray}{59.5} & \textcolor{gray}{51.2} \\
        Detic \cite{detic} & Caption   & RN50 & 27.8 & \textcolor{gray}{51.1} & \textcolor{gray}{45.0} \\
        OV-DETR$^\dagger$ \cite{ovdetr} & CLIP & RN50     & 29.4 & \textcolor{gray}{61.0} & \textcolor{gray}{52.7} \\
        ProxyDet \cite{proxydet} & Caption & RN50     & 30.4 & \textcolor{gray}{52.6} & \textcolor{gray}{46.8} \\
        RegionCLIP \cite{regionclip} & Caption    & RN50         & 31.4 & \textcolor{gray}{57.1} & \textcolor{gray}{50.4} \\
        RTGen \cite{rtgen} & Caption   & RN50 & 33.6 & \textcolor{gray}{51.7} & \textcolor{gray}{46.9} \\
        BARON-KD \cite{wu2023aligning} & CLIP   & RN50     & 34.0 & \textcolor{gray}{60.4} & \textcolor{gray}{53.5} \\
        CLIM \cite{clim} &  CLIP  & RN50     & 36.9 & \textcolor{gray}{-} & \textcolor{gray}{-} \\
        SAS-Det \cite{sasdet} & CLIP    & RN50 & 37.4 & \textcolor{gray}{58.5} & \textcolor{gray}{53.0} \\
        RegionCLIP \cite{regionclip} & Captions  & RN50x4       & 39.3 & \textcolor{gray}{61.6} & \textcolor{gray}{55.7} \\
        CORA$^\dagger$ \cite{wu2023cora} & CLIP   & RN50x4    & 41.7 & \textcolor{gray}{44.5} & \textcolor{gray}{43.8} \\
        OV-DQUO$^\dagger$ \cite{ovdquo} & CLIP     & RN50x4       & 45.6 & \textcolor{gray}{-} & \textcolor{gray}{-} \\
        \midrule
        RO-ViT \cite{kim2023region} &  CLIP   & ViT-L/16     & 33.0 & \textcolor{gray}{-} & \textcolor{gray}{47.7} \\
        CFM-ViT \cite{CFM} & CLIP   & ViT-L/16     & 34.1 & \textcolor{gray}{-} & \textcolor{gray}{46.0} \\
        F-ViT \cite{wu2023clipself} & CLIP  & ViT-B/16  & 37.6 & \textcolor{gray}{54.9} & \textcolor{gray}{50.4} \\
        BIND \cite{bind} &CLIP & ViT-L/16  & 41.5 & \textcolor{gray}{58.3} & \textcolor{gray}{54.8} \\
        F-ViT \cite{wu2023clipself} & CLIP  & ViT-L/14  & 44.3 & \textcolor{gray}{64.1} & \textcolor{gray}{59.0} \\
        \midrule
        \rowcolor[HTML]{EFEFEF}F-ViT+DeCLIP & CLIP     & ViT-B/16       &41.1  & \textcolor{gray}{57.8} & \textcolor{gray}{53.5} \\
        \rowcolor[HTML]{EFEFEF}F-ViT+DeCLIP & CLIP     & ViT-L/14       &46.2  & \textcolor{gray}{65.2} & \textcolor{gray}{60.3} \\
        \rowcolor[HTML]{EFEFEF}OV-DQUO+DeCLIP$^\dagger$ & CLIP     & ViT-B/16       &46.1  & \textcolor{gray}{56.3} & \textcolor{gray}{53.6} \\
        \rowcolor[HTML]{EFEFEF}OV-DQUO+DeCLIP$^\dagger$ & CLIP     & ViT-L/14       & \textbf{48.3} & \textcolor{gray}{60.0} & \textcolor{gray}{56.9} \\
        \bottomrule
      \end{tabular}
    \end{adjustbox}
  \end{subtable}
  \hspace{10mm}
\begin{subtable}[t]{0.462\textwidth}
  \centering
    \caption{OV-LVIS benchmark \cite{lvis}}  
    \begin{adjustbox}{width=\linewidth,center,valign=t}
      \begin{tabular}{l|l|l|cccc}
        \toprule
        Method & Supervision & Backbone & $\text{mAP}_{r}$ & \textcolor{gray}{$\text{mAP}_{c}$} & \textcolor{gray}{$\text{mAP}_{f}$} & \textcolor{gray}{$\text{mAP}$} \\
        \midrule
        ViLD \cite{vild} &CLIP & RN50 & 16.6 & \textcolor{gray}{24.6} & \textcolor{gray}{30.3} & \textcolor{gray}{25.5} \\
        OV-DETR$^\dagger$ \cite{ovdetr} &CLIP & RN50 & 17.4 & \textcolor{gray}{25.0} & \textcolor{gray}{32.5} & \textcolor{gray}{26.6} \\
        BARON-KD \cite{wu2023aligning} &CLIP & RN50 & 22.6 & \textcolor{gray}{27.6} & \textcolor{gray}{29.8} & \textcolor{gray}{27.6} \\
        RegionCLIP \cite{regionclip} &Caption & RN50x4 & 22.0 & \textcolor{gray}{32.1} & \textcolor{gray}{36.9} & \textcolor{gray}{32.3} \\
        CORA$^{+}$$^\dagger$ \cite{wu2023cora} &Caption & RN50x4 & 28.1 & \textcolor{gray}{-} & \textcolor{gray}{-} & \textcolor{gray}{-} \\
        SAS-Det \cite{sasdet} &CLIP & RN50x4 & 29.1 & \textcolor{gray}{32.4} & \textcolor{gray}{36.8} & \textcolor{gray}{33.5} \\
        CLIM \cite{clim} &  CLIP  & RN50x64     & 32.3 & \textcolor{gray}{-} & \textcolor{gray}{-} & \textcolor{gray}{-} \\
        F-VLM \cite{fvlm} &CLIP & RN50x64 & 32.8 & \textcolor{gray}{-} & \textcolor{gray}{-} & \textcolor{gray}{34.9} \\
        \midrule
        F-ViT \cite{wu2023clipself} &CLIP & ViT-B/16 & 25.3 & \textcolor{gray}{21.8} & \textcolor{gray}{29.1} & \textcolor{gray}{25.2} \\
        RTGen \cite{rtgen} & Caption   & Swin-B & 30.2 & \textcolor{gray}{39.9} & \textcolor{gray}{41.3} & \textcolor{gray}{38.8} \\
        BIND \cite{bind} &CLIP & ViT-L/16 & 32.5 & \textcolor{gray}{33.4} & \textcolor{gray}{35.3} & \textcolor{gray}{33.2} \\
        Detic \cite{detic} & Caption   & Swin-B & 33.8 & \textcolor{gray}{-} & \textcolor{gray}{-} & \textcolor{gray}{47.0} \\
        CFM-ViT \cite{CFM} &CLIP & ViT-L/14 & 33.9 & \textcolor{gray}{-} & \textcolor{gray}{-} & \textcolor{gray}{36.6} \\
        RO-ViT \cite{kim2023region} &CLIP & ViT-H/16 & 34.1 & \textcolor{gray}{-} & \textcolor{gray}{-} & \textcolor{gray}{35.1} \\
        F-ViT \cite{wu2023clipself} &CLIP & ViT-L/14 & 34.9 & \textcolor{gray}{34.6} & \textcolor{gray}{35.6} & \textcolor{gray}{35.1} \\
        ProxyDet \cite{proxydet} &Caption & Swin-B & 36.7 & \textcolor{gray}{-} & \textcolor{gray}{-} & \textcolor{gray}{41.5} \\
        CoDet \cite{codet} &Caption & ViT-L/14 & 37.0 & \textcolor{gray}{46.3} & \textcolor{gray}{46.3} & \textcolor{gray}{44.7} \\
        OV-DQUO$^\dagger$ \cite{ovdquo} &CLIP & ViT-L/14 & 39.3 & \textcolor{gray}{-} & \textcolor{gray}{-} & \textcolor{gray}{-} \\
        \midrule
        \rowcolor[HTML]{EFEFEF}F-ViT+DeCLIP & CLIP     & ViT-B/16       & 26.8  & \textcolor{gray}{22.4} & \textcolor{gray}{29.8} & \textcolor{gray}{26.0} \\
        \rowcolor[HTML]{EFEFEF}F-ViT+DeCLIP & CLIP     & ViT-L/14       & 37.2 & \textcolor{gray}{35.2} & \textcolor{gray}{36.5} & \textcolor{gray}{36.0} \\
        \rowcolor[HTML]{EFEFEF}OV-DQUO+DeCLIP$^\dagger$ & CLIP     & ViT-B/16       &31.0    & \textcolor{gray}{-} & \textcolor{gray}{-} & \textcolor{gray}{27.7} \\
        \rowcolor[HTML]{EFEFEF}OV-DQUO+DeCLIP$^\dagger$ & CLIP     & ViT-L/14       & \textbf{41.5}   & \textcolor{gray}{-} & \textcolor{gray}{-} & \textcolor{gray}{34.6} \\
        \bottomrule
      \end{tabular}
    \end{adjustbox}
  \end{subtable}
\label{app_tab4}
\end{table*}
\subsection{Analyses of Feature Correlations}
\label{corelation}
We have analyzed CLIP and found that its limitation in open-vocabulary dense prediction arises from image tokens failing to aggregate information from spatially or semantically related regions. Figure~\ref{app_fig2} presents a comparison of feature correlations among CLIP~\cite{clip}, DeCLIP, and existing pre-finetuning methods~\cite{wu2023clipself,regionclip} at each vision encoder layer. 
\par This experiment provide insight into how the output features of each layer in CLIP's visual encoder changed after fine-tuning. In this experiment, we randomly select an image token from the primary object within the image (i.e., the bird) as the anchor and visualize the cosine similarity between the anchor and the other image tokens. 
The experimental results indicate that the impact of various fine-tuning methods on the correlation of CLIP's output features becomes noticeable starting from the 6th encoder layer.

\mypara{CLIP vs. existing pre-fine-tuning methods.} Rows 1, 2, and 3 of Figure~\ref{app_fig2} exhibit the changes in feature correlations of CLIP after region-level fine-tuning~\cite{regionclip,wu2023clipself}. The experimental results indicate that region-level fine-tuning enhances the feature correlations of the anchor image token to start converging towards the object it belongs to (rows 2 and 3), rather than being randomly scattered across the image (row 1). 
\par This change is highly effective for open-vocabulary object detection tasks. As relevant features become more focused, region features exhibit enhanced discriminative power in the visual-language space when extracting the object's region features from the image for recognition. However, these methods remain constrained in image segmentation tasks that demand pixel-level precision. As shown in the feature correlation results in rows 2 and 3 of Figure~\ref{app_fig2}, most of the pixels surrounding the bird will be misclassified as “bird” rather than to be “background”.

\mypara{CLIP vs. DeCLIP.} Rows 1 and 4 of Figure~\ref{app_fig2} exhibit the changes in feature correlations of CLIP after decoupled feature enhancement strategy. The experimental results indicate that DeCLIP enhances the feature correlations of the anchor image token to closely align with the object it represents, in clear contrast with other existing pre-fine-tuning approaches (row 2 and 3). This experiment reveals why DeCLIP is better suited for image segmentation tasks than existing methods. Additionally, the experiment demonstrates DeCLIP's also superiority over current pre-finetuning approaches in region classification tasks. As shown in the feature correlation map of DeCLIP's 12th layer, the image regions corresponding to the same object as the anchor image token display a strong red color, indicating a very high feature correlation strength in these regions, thereby enhancing the discriminative power of region features within the visual-language space.

\begin{table}[tbp]
\renewcommand{\arraystretch}{0.9}
\centering
\caption{Detailed comparison of transferring LVIS-trained detectors to the COCO and Objects365 datasets.}
\begin{adjustbox}{width=\linewidth,center,valign=t}
\begin{tabular}{l|ccc|cccccc}
\toprule 
\multirow{2.5}{*}{Method} & \multicolumn{3}{c|}{COCO \cite{mscoco}} & \multicolumn{6}{c}{Objects365 \cite{object365}} \\
\cmidrule(lr){2-4} \cmidrule(lr){5-10} 
& AP & AP$_{50}$ & AP$_{75}$ & AP & AP$_{50}$ & AP$_{75}$ & AP$_s$ & AP$_m$ & AP$_l$ \\
\midrule
Supervised Baseline \cite{vild} & 46.5 & 67.6 & 50.9 & 25.6 & 38.6 & 28.0 & - & - & - \\
\midrule
ViLD \cite{vild} & 36.6 & 55.6 & 39.6 & 11.8 & 18.0 & 12.6 & - & - & - \\
DetPro \cite{du2022learning} & 34.9 & 53.8 & 37.4 & 12.1 & 18.8 & 12.9 & 4.5 & 11.5 & 18.6 \\
BARON \cite{wu2023aligning} & 36.2 & 55.7 & 39.1 & 13.6 & 21.0 & 14.5 & 5.0 & 13.1 & 20.7 \\
F-VLM \cite{fvlm} & 37.9 & 59.6 & 41.2 & 16.2 & 25.3 & 17.5 & - & - & - \\
CoDet \cite{codet} & 39.1 & 57.0 & 42.3 & 14.2 & 20.5 & 15.3 & - & - & - \\
RO-ViT \cite{rovit} & - & - & - & 17.7 & 27.4 & 19.1 & - & - & - \\
CLIPSelf \cite{wu2023clipself} & 40.5 & 63.8 & 44.3 & 19.5 & 31.3 & 20.7 & 9.7 & 23.2 & 35.5 \\
\midrule
 \rowcolor[HTML]{EFEFEF}\textbf{DeCLIP} & \textbf{41.0} & \textbf{64.6} & \textbf{44.8} & \textbf{20.0} & \textbf{32.2} & \textbf{21.2} & \textbf{10.0} & \textbf{24.4} & \textbf{36.7} \\
\bottomrule
\end{tabular}
\end{adjustbox}
\label{app_tab5}
\end{table}
\subsection{Comparison of Semantic Segmentation Results}
\label{semantic_result}
\begin{figure*}[tbp]
  \centering
  \includegraphics[width=.95\textwidth]{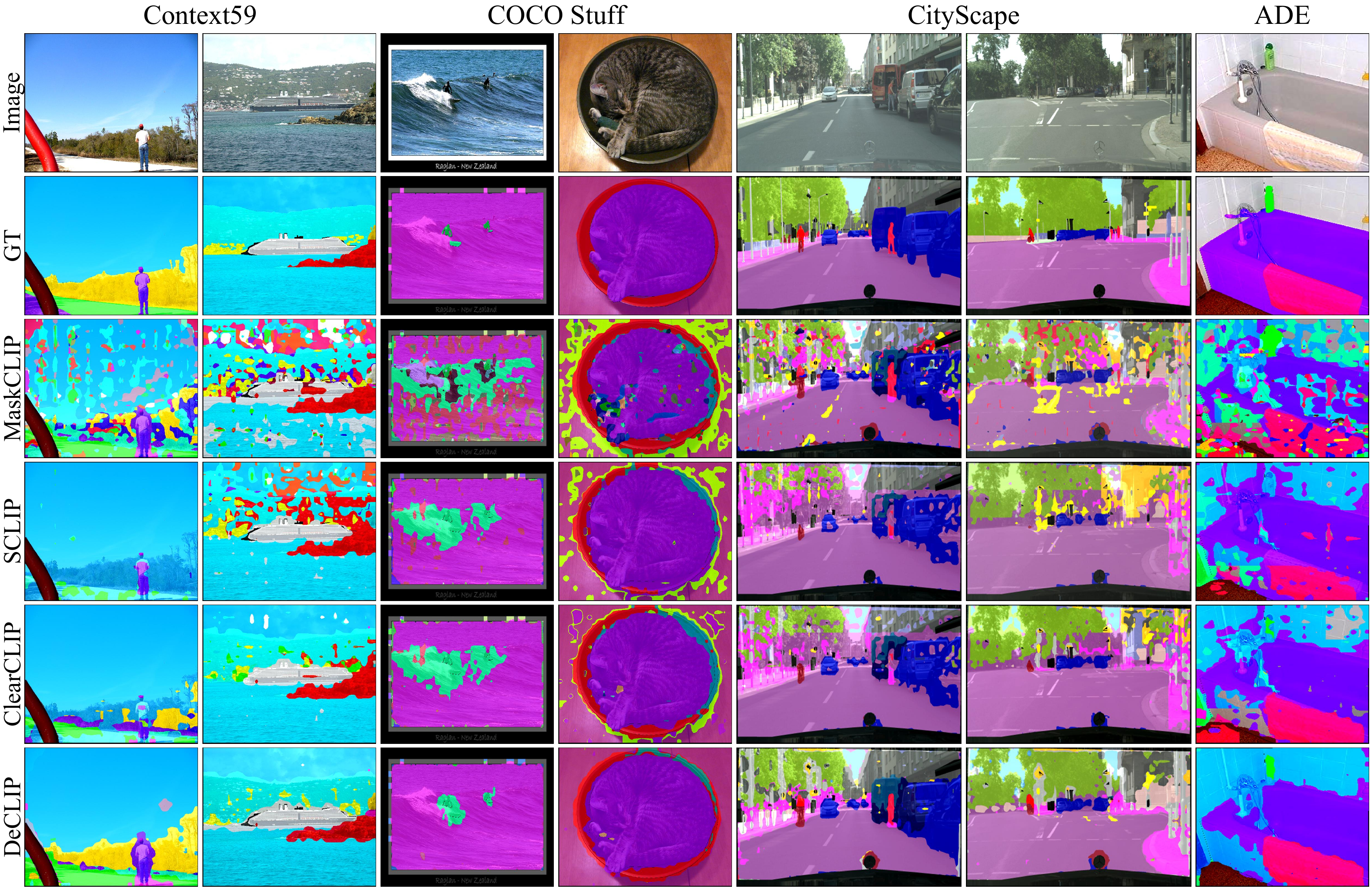}
  \caption{Qualitative comparison of the open-vocabulary semantic segmentation results between DeCLIP and existing approaches~\cite{maskclip,sclip,clearclip}.}
  \label{app_fig3}
\end{figure*}

Figure~\ref{app_fig3} shows a qualitative comparison of MaskCLIP~\cite{maskclip}, SCLIP~\cite{sclip}, ClearCLIP~\cite{clearclip}, and our proposed DeCLIP across the Context59~\cite{context}, COCO-Stuff~\cite{cocostuff}, Cityscapes~\cite{cityscape}, and ADE20K~\cite{ade20k} datasets. We observe that, compared to other methods, DeCLIP consistently produces higher-quality and more precise segmentation maps. 

\par Specifically, benefiting from content feature distillation, which improves the discriminability of local features, DeCLIP successfully recognizes trees, people, and curbs in the images, as shown in columns 1, 5, and 6 of Figure~\ref{app_fig3}, whereas other models fail. Furthermore, our observation indicates that the distillation of context features improves the spatial consistency of DeCLIP's local features, leading to smoother and less noisy segmentation results compared to other models, as demonstrated in columns 2, 3, 4, and 7 of Figure~\ref{app_fig3}. This demonstrates the superiority of our decoupled feature enhancement strategy. 
\subsection{Comparison of Attention Maps}
\label{attention_map}
\begin{figure*}[tbp]
  \centering
  \includegraphics[width=\textwidth]{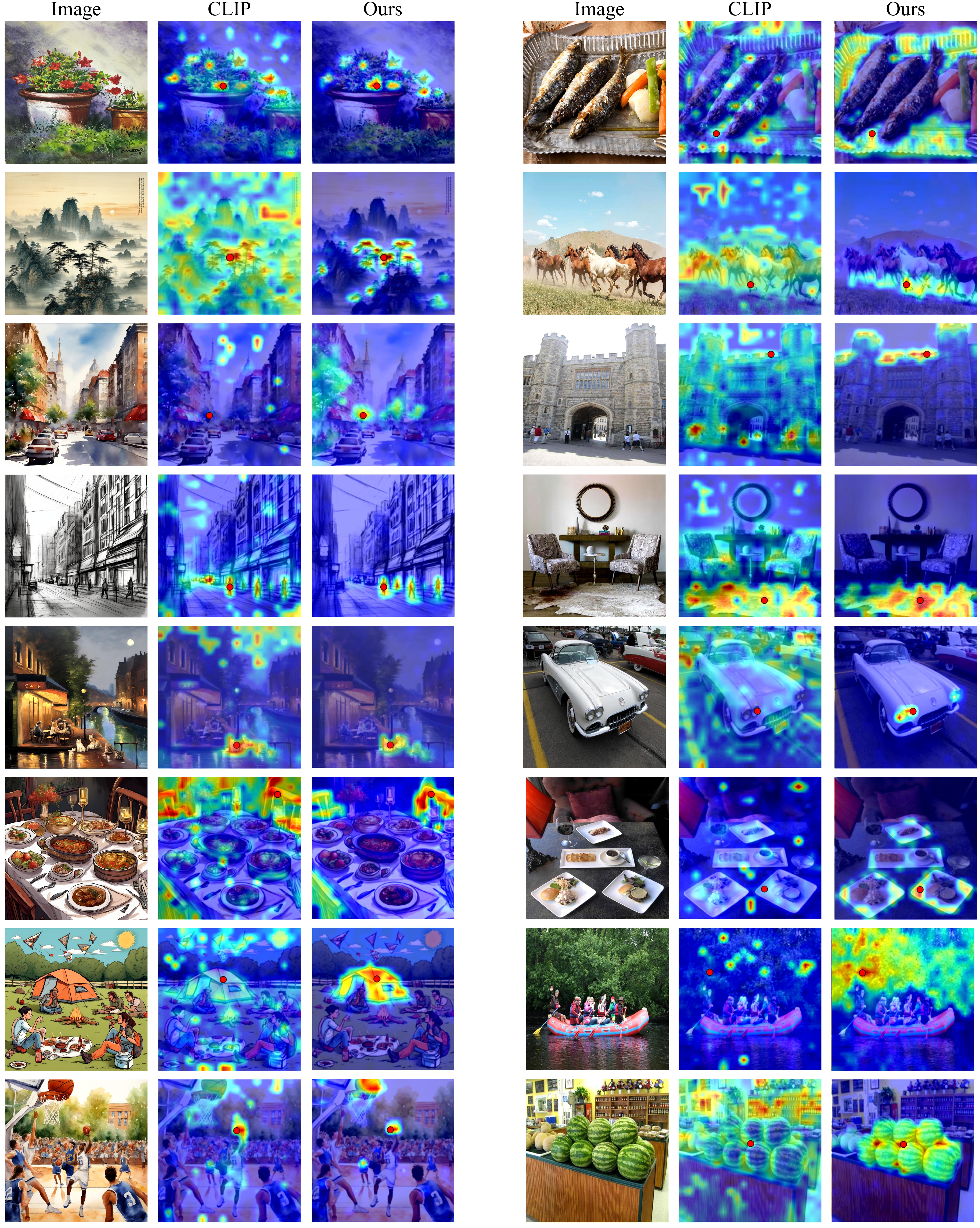}
  \caption{Comprehensive comparison of attention maps between CLIP and DeCLIP. The left side presents images of various styles generated by generative models~\cite{diffusion}. The images presented on the right-hand side comes from a subset of images in the Object365~\cite{object365} validation set. Anchor image token marked in red.}
  \label{app_fig4}
\end{figure*}

Figure~\ref{app_fig4} offers a detailed comparison of attention maps between CLIP and our proposed DeCLIP approach. As DeCLIP involves unsupervised fine-tuning, we conducted tests using diverse cross-domain image styles to thoroughly assess its generalization capability. Specifically, we utilized generative models~\cite{diffusion} to generate test images in various styles such as ink painting, watercolor, sketch, animation, and oil painting, which are depicted on the left side of Figure~\ref{app_fig4}. These cross-domain test images were not part of the fine-tuning dataset for DeCLIP (i.e., COCO2017~\cite{mscoco}).
\par In addition, we performed a detailed comparison of attention maps between CLIP and DeCLIP on in-domain images. Specifically, we selected a subset of images from the Object365~\cite{object365} validation set for testing, with the results shown on the right-hand side of Figure ~\ref{app_fig4}. During the testing phase, we first resized the images to 336×336 pixels and then fed them into the model to extract features. Subsequently, we randomly selected an anchor image token and visualized its attention map in the 12th attention block, as indicated by the red dots on the test images in Figure~\ref{app_fig4}. For details on the calculation process of the attention map, please refer to Figure~\ref{app_fig1}.
\par As depicted in Figure~\ref{app_fig4} , due to the proxy token phenomenon, the heatmap generated by the anchor image token in vanilla CLIP frequently lacks semantic consistency with its corresponding object. In contrast, despite being fine-tuned only on the natural scene dataset COCO, DeCLIP demonstrates significant semantic relevance for both in-domain and cross-domain test images. Moreover, benefiting from context feature distillation, DeCLIP's semantic correlations demonstrate remarkably fine granularity, effectively outlining the boundaries of each object semantically associated with the anchor image token.

\section{Details of Experimental Settings}
\label{exp_settings}
In this section, we present further details and configurations utilized in our experiments.
\subsection{Datasets and Evaluation Protocols}
\label{dataset}
\mypara{Open-Vocabulary Detection.} Following established settings~\cite{ovr-cnn,wu2023clipself,wu2023cora}, we evaluated our model on the OV-COCO \cite{mscoco}, OV-LVIS \cite{lvis}, COCO, and Object365 \cite{object365} datasets. The OV-COCO dataset includes 48 base categories and 17 novel categories. The training set contains only base categories, totaling 107,761 images, while the validation set comprises 4,836 images featuring both base and novel categories. We report the mean Average Precision (mAP) at an Intersection over Union (IoU) threshold of 0.5 for novel categories. The OV-LVIS dataset consists of 1,203 categories. Its training set includes only 461 common and 405 frequent categories, totaling 100,170 images. The validation set contains 19,809 images with common, frequent, and rare categories. We report the mAP for rare categories at IoU thresholds ranging from 0.5 to 0.95. Additionally, we provide cross-dataset evaluation results on the COCO and Object365 validation sets for models trained on OV-LVIS to assess generalization across domains.

\mypara{Open-Vocabulary Semantic Segmentation.} In line with prior studies \cite{catseg}, we trained our model on the COCO-Stuff dataset \cite{cocostuff}, which comprises 118,000 images with dense annotations across 171 categories. We then evaluated the model on the ADE20K \cite{ade20k}, PASCAL VOC \cite{voc}, and PASCAL-Context \cite{context} datasets. ADE20K \cite{ade20k} includes 20,000 training images and 2,000 validation images, with two category sets: A-150 (150 common categories) and A-847 (847 categories) \cite{ade847}. PASCAL-Context consists of 5,000 training and validation images, with category sets PC-59 (59 categories) and PC-459 (459 categories). The PASCAL VOC dataset includes 1,500 images for training and validation, featuring category sets PAS-20 (20 categories) and PAS-21 (20 object categories plus one background class). We used mean Intersection over Union (mIoU) as the evaluation metric in all experiments.

\mypara{Open-Vocabulary Semantic Segmentation Based on VLM Features.} To further evaluate DeCLIP, we assessed it on six commonly used semantic segmentation benchmarks: PASCAL VOC 2012 \cite{voc}, PASCAL Context \cite{context}, Cityscapes \cite{cityscape}, ADE20K \cite{ade20k}, COCO-Stuff \cite{mscoco}, and COCO-Object \cite{cocostuff}. For datasets including a background category, we refer to them as VOC21 and Context60; those without a background category are termed VOC20 and Context59. Consistent with previous experiments, we used mIoU as the evaluation metric across these benchmarks. 

\subsection{Implementation Details}
\label{detail}

\mypara{DeCLIP.} DeCLIP was trained on training set images from the COCO2017~ \cite{mscoco} dataset using 8 GPUs, each with a batch size of 2, for 6 epochs (about 44 min/epoch on 8×4090 GPUs). The AdamW~\cite{adamw}  optimizer with a learning rate of $1\mathrm{e}{-5}$ and a weight decay of 0.1 was employed during the training process.

\par During the content feature distillation process, the image is divided into $k$ blocks, where $k = m \times n$, and $m$ and $n$ are randomly sampled from the range [1, 6]. After cropping $k$ image blocks from the original image, the patches are resized to a resolution of 224×224 and subsequently fed into the teacher model to generate the corresponding [CLS] tokens for content feature distillation. Unless stated otherwise, our experiments were conducted using EVA-CLIP~\cite{evaclip}.
\par In the process of context feature distillation, given the distinct image preprocessing methods with varying means and standard deviations used by CLIP and VFM during pretraining, we incorporated the corresponding parameters during the distillation process. Additionally, to address the potential variation in patch sizes between CLIP and VFM (e.g., CLIP uses a 16-patch size while DINOV2 uses a 14-patch size), we adjusted the image resolutions to maintain consistency in the number of image tokens. For example, we set the resolution of CLIP to 1024 and that of DINOV2 to 896, ensuring both models possess 4096 image tokens. The weight $\lambda$ for context feature distillation is established at 0.25. Unless specified otherwise, our default VFM is DINOv2~\cite{dinov2}.  

\mypara{Open-vocabulary detection.} In the open-vocabulary detection experiment, DeCLIP was evaluated in two model baselines: F-ViT~\cite{wu2023clipself} and OV-DQUO~\cite{ovdquo}. These baselines are constructed based on transfer learning principles, utilizing the image encoder of CLIP for feature extraction while maintaining the backbone network frozen during training and only  training the task-specific components. The two baseline models utilize distinct detector architectures: F-ViT employs the traditional Faster R-CNN~\cite{fasterrcnn} architecture, whereas OV-DQUO utilizes the modern Detection Transformer~\cite{detr} architecture. This enables a thorough assessment of the efficacy of our proposed approach.

\par We maintained the default training strategies and hyperparameter configurations from the original studies for both baseline models to uphold experiment fairness. The only modification was to the temperature parameter when integrating DeCLIP for object detection. For F-ViT, the temperature was set to 45 for the OV-COCO benchmark and 90 for the OV-LVIS benchmark. In OV-DQUO, the temperature was set to 50 for both the OV-COCO and OV-LVIS benchmarks.

\mypara{Open-Vocabulary Semantic Segmentation.} In the open-vocabulary semantic segmentation experiments, we applied DeCLIP to the CAT-Seg \cite{catseg} baseline. For all experiments, we adhered to the default training and inference settings of vanilla CAT-Seg, replacing only the image encoder with DeCLIP. 

\mypara{Open-Vocabulary Semantic Segmentation Based on VLM Features.} During inference, we resized the shorter side of images to 448 pixels and employed a sliding window strategy with a window size of 336×336 and a stride of 112×112. For all datasets, we generate textual descriptions by utilizing the standard ImageNet prompts~\cite{clip} in conjunction with their respective class names. No post-processing steps were applied. 
\section{Related Work}
\label{related_work}
\subsection{Open-Vocabulary Dense Prediction}
\label{dense_prediction}
Open-vocabulary dense prediction aims to detect and segment visual concepts from novel categories using textual descriptions, extending beyond the base categories on which the model was trained. According to recent surveys~\cite{ovdsurvey}, methods in this field can be broadly classified into four categories: knowledge distillation-based~\cite{oadp,wu2023aligning,ovdetr,gkc}, pseudo-labeling~\cite{regionclip,sasdet,detic,ovdquo,ttd}, region-aware training~\cite{wu2023cora,rovit,CFM,groupvit,openseg}, and transfer learning-based approaches~\cite{fvlm,ovdquo,wu2023clipself,lseg,ZegFormer,OVSeg,ovdiff}.
\par Knowledge distillation-based methods, such as ViLD \cite{vild}, BARON \cite{wu2023aligning}, and OADP \cite{oadp}, propose various distillation frameworks to transfer the generalized classification knowledge of VLMs \cite{clip,evaclip} into dense prediction models. Pseudo-labeling methods like RegionCLIP \cite{regionclip} and SAS-Det~\cite{sasdet} enhance region-text alignment by generating pseudo-labels for image-text pairs using VLMs or self-training techniques. Region-Aware Training methods, exemplified by CORA~\cite{wu2023cora}, improve the object classification accuracy of CLIP by learning region prompts.
\par Transfer Learning-Based methods~\cite{ovdquo,wu2023clipself,catseg,fcclip,maskqclip,MAFT,maft+,ZegFormer,OVSeg,ovdiff} utilize the image encoder of VLM as a feature extractor and exclusively train lightweight task-specific components. These methods have become mainstream in open-vocabulary dense prediction due to their broad applicability. While leveraging VLMs as feature extractors offers significant advantages due to their comprehensive pre-training, directly applying these image-level models to dense prediction tasks often results in domain shift issues \cite{wu2023cora,wu2023clipself}, thereby limiting their performance. In this paper, we integrate DeCLIP into transfer learning-based object detection baselines F-ViT and OV-DQUO, as well as the image segmentation baseline CATSeg, to enhance their performance in open-vocabulary dense prediction tasks.

% Several studies have attempted to alleviate the domain shift issue in applying CLIP to dense prediction tasks via fine-tuning strategies. These approaches fall into two main categories:
% These methods fine-tune CLIP while training task-specific components \cite{MAFT,catseg,lseg,OVSeg,maskqclip,maft+,xie2024sed}. For instance, CAT-Seg \cite{catseg} proposes an attention fine-tuning strategy based on ViT CLIP, which generalizes well to unseen categories. MAFT \cite{MAFT} leverages attention bias to fine-tune CLIP for mask classification. 
% \item \mypara{Pre-fine-tuning.} These methods directly fine-tune CLIP using cost-efficient techniques \cite{wu2023clipself,pacl,regionclip,clim,wu2023cora}, which are more closely aligned with the approach proposed in this paper. As illustrated in Figure \ref{app_fig4}(a), CLIM \cite{clim} employs a mosaic augmentation technique to stitch multiple images into a single image, enabling each sub-image to serve as a pseudo-region for region-text contrastive learning. CLIPSelf \cite{wu2023clipself} enhances CLIP's region classification accuracy by maximizing cosine similarity between its region representations and the corresponding image crop representations, as illustrated in Figure \ref{app_fig4}(b).

\subsection{Transferring VLMs to Dense Prediction Tasks}
\label{transfer}
As VLMs~\cite{clip,evaclip} were initially trained on image-text pairs, the direct application of these image-level models to dense prediction tasks, which require region-level or pixel-level semantic understanding, results in significant performance degradation. Several studies have attempted to address this limitation through fine-tuning strategies. These approaches can be broadly categorized into joint fine-tuning and pre-fine-tuning approaches.

\par Joint fine-tuning methods fine-tune CLIP while training task-specific components \cite{MAFT,catseg,lseg,OVSeg,maskqclip,maft+,xie2024sed}. For instance, CAT-Seg \cite{catseg} proposes an attention fine-tuning strategy based on ViT CLIP, which generalizes well to unseen categories. MAFT \cite{MAFT} leverages attention bias to fine-tune CLIP for mask classification. 

\par Pre-fine-tuning methods directly fine-tune CLIP using cost-efficient techniques \cite{wu2023clipself,pacl,regionclip,clim,wu2023cora}. For instance, CLIM \cite{clim} employs a mosaic augmentation technique to stitch multiple images into a single image, enabling each sub-image to serve as a pseudo-region for region-text contrastive learning. CLIPSelf \cite{wu2023clipself} enhances CLIP's region classification accuracy by maximizing cosine similarity between its region representations and the corresponding image crop representations. 
\par Despite the promising results of the two categories of fine-tuned methods, they continue to exhibit certain limitations. In contrast to these studies, we conduct an analysis of CLIP and identify that its limitation in open-vocabulary dense prediction stems from the inability of image tokens to effectively aggregate information from spatially or semantically related regions. To address this, we propose integrating VFMs into the pre-fine-tuning process and decoupling features for distillation, thereby improving the discriminability and spatial consistency of CLIP’s local features.
\subsection{Vision Foundation Models}
\label{vfm}
Vision foundation models, including the Self-Supervised Representation Learning (SSL) series~\cite{dino,dinov2,ibot,mim,moco,mlcd} and the SAM series~\cite{sam,sam2}, which are trained on large-scale segmentation data, demonstrate the ability to extract features that exhibit strong spatial consistency.
\par SSL is a key area in computer vision that focuses on learning meaningful visual features without manual annotations~\cite{dino,dinov2,ibot,mim,moco,mlcd}. Vision models trained through SSL can extract image features with excellent spatial understanding. For example, the DINO series~\cite{dino,dinov2} can identify similar semantic regions across different images and segment main objects without explicit supervision. Another prominent vision foundation model is SAM~\cite{sam,sam2}, which demonstrates similarly outstanding spatial understanding. Trained on the extensive SA-1B segmentation dataset, SAM can accurately capture and segment objects regions in images based on prompts.
\par Recently, some studies have explored the combination of CLIP with VFM, such as SAM-CLIP~\cite{samclip}, OV-SAM~\cite{ovsam}, and FrozenSeg~\cite{frozenseg}, with the goal of integrating SAM's powerful image segmentation capabilities and CLIP's zero-shot semantic perception capabilities. AM-RADIO~\cite{am_radio} trains a unified vision model through multi-teacher distillation from multiple foundational vision models such as CLIP, DINOv2, and SAM. However, SAM-CLIP, OV-SAM, and FrozenSeg focus on integrating CLIP into SAM rather than enhancing CLIP itself as DeCLIP does. AM-RADIO does not support OVSS, as confirmed by its authors in Github issues (No.~81, 55, and 42). Another study that solves similar problems to DeCLIP is ViT-Register~\cite{register}. However, unlike DeCLIP, ViT-Register~\cite{register} does not solve the dense perception deficiency arising from CLIP's image-text alignment.